\documentclass{article}

     \PassOptionsToPackage{numbers, sort&compress}{natbib}
     \usepackage[final]{neurips_2020}

\usepackage[utf8]{inputenc} % allow utf-8 input
\usepackage[T1]{fontenc}    % use 8-bit T1 fonts
\usepackage{hyperref}       % hyperlinks
\usepackage{url}            % simple URL typesetting
\usepackage{booktabs}       % professional-quality tables
\usepackage{amsfonts}       % blackboard math symbols
\usepackage{nicefrac}       % compact symbols for 1/2, etc.
\usepackage{microtype}      % microtypography

\usepackage{amsmath}
\usepackage{amssymb} % \mathbb
\usepackage{amsthm}
\usepackage{stmaryrd} % \llbracket
\usepackage{dsfont} % \mathds{..}
\usepackage{paralist}
\usepackage{cleveref} % \cref{..}  %% [nameinlink]
\usepackage{accents}
\usepackage{xifthen}
\usepackage{xcolor} % \definecolor{..}
\usepackage{enumitem}
\usepackage[pdftex]{graphicx}
\usepackage{subcaption}
\newtheorem{definition}{Definition}
\newtheorem{theorem}{Theorem}
\newtheorem{lemma}[theorem]{Lemma}
\newtheorem{proposition}[theorem]{Proposition}
\newtheorem{corollary}[theorem]{Corollary}

\newtheorem{remark}{Remark}

\crefname{definition}{Definition}{Definitions}
\crefname{theorem}{Theorem}{Theorems}
\crefname{lemma}{Lemma}{Lemmas}
\crefname{proposition}{Proposition}{Propositions}
\crefname{corollary}{Corollary}{Corollarys}
\crefname{problem}{Problem}{Problems}
\crefname{example}{Example}{Examples}
\crefname{fact}{Fact}{Facts}
\crefname{conjecture}{Conjecture}{Conjectures}
\crefname{remark}{Remark}{Remarks}
\crefname{condition}{Condition}{Conditions}
\crefname{requirement}{Requirement}{Requirements}
\crefname{figure}{Figure}{Figures}

\newtheoremstyle{myproofsty}% name of the style to be used
{\topsep}% measure of space to leave above the theorem. E.g.: 3pt
{\topsep}% measure of space to leave below the theorem. E.g.: 3pt
{\normalfont}% name of font to use in the body of the theorem
{0pt}% measure of space to indent
{\itshape}% name of head font
{}% punctuation between head and body
{5.5pt}% space after theorem head; " " = normal interword space
{\thmname{#1}\ifthenelse{\isempty{#3}}{}{ of {#3}}.}
\theoremstyle{myproofsty}
\newtheorem{myproof}{Proof}

\makeatletter
\newsavebox{\@brx}
\newcommand{\llangle}[1][]{\savebox{\@brx}{\(\m@th{#1\langle}\)}%
  \mathopen{\copy\@brx\mkern2mu\kern-0.9\wd\@brx\usebox{\@brx}}}
\newcommand{\rrangle}[1][]{\savebox{\@brx}{\(\m@th{#1\rangle}\)}%
  \mathclose{\copy\@brx\mkern2mu\kern-0.9\wd\@brx\usebox{\@brx}}}
\makeatother

\DeclareMathSymbol{\widetildesym}{\mathord}{largesymbols}{"65}
\newcommand\lowerwidetildesym{%
  \text{\smash[b]{\raisebox{-1.45ex}{%
        $\widetildesym$}}}}
\newcommand\mytilde[1]{%
  \mathchoice
      {\accentset{\displaystyle\lowerwidetildesym}{#1}}
      {\accentset{\textstyle\lowerwidetildesym}{#1}}
      {\accentset{\scriptstyle\lowerwidetildesym}{#1}}
      {\accentset{\scriptscriptstyle\lowerwidetildesym}{#1}}
}

\renewcommand\bar[1]{\accentset{\rule{.4em}{.8pt}}{#1}}

\newcommand*{\bR}{\mathbb{R}}
\newcommand*{\bZ}{\mathbb{Z}}
\newcommand*{\cX}{\mathcal{X}}
\newcommand*{\cY}{\mathcal{Y}}
\newcommand*{\cZ}{\mathcal{Z}}

\newcommand*{\indc}[1]{{\mathds{1}[{#1}]}}

\newcommand*{\sem}[1]{\llbracket{#1}\rrbracket}
\newcommand*{\semdiff}[1]{\llbracket{#1}\rrbracket^{\nabla}\;\!\!} %% \;\!
\newcommand*{\eval}[1]{\llangle{#1}\rrangle} %% {\langle{#1}\rangle}
\newcommand*{\ideriv}{\partial_\bullet} %%{\partial_I} {\eth}
\newcommand*{\iderivk}[1]{\ideriv^{{#1}}}
\newcommand*{\irepr}[1]{\Gamma({#1})}

\newcommand*{\intrr}[1]{{\rm int}({#1})}
\newcommand*{\subintrr}[1]{{\rm subint}({#1})}
\newcommand*{\refine}[1]{{\rm refine}({#1})}
\newcommand*{\ifelse}[3]{{\rm cond}({#1},{#2},{#3})}

\newcommand*{\code}[1]{\mathtt{#1}}
\newcommand*{\codett}[1]{\mathtt{#1}}
\newcommand*{\AD}{\mathrm{AD}}

\newcommand*{\Remove}{\mathrm{REM}}
\newcommand*{\Zero}{\mathrm{ZERO}}
\newcommand*{\One}{\mathrm{ONE}}

\newcommand*{\Merge}{\mathrm{MERGE}}
\newcommand*{\Map}{\mathrm{MAP}}
\newcommand*{\PrimD}{\mathrm{PrimD}}
\newcommand*{\RecT}{\mathrm{RecT}}

\newcommand{\cmt}[2]{{\color{blue}{\bf [}{\bf #1:} {\it #2}{\bf ]}}}

\newcommand{\hsy}[1]{\cmt{HSY}{#1}}
\newcommand{\wl}[1]{\cmt{WL}{#1}}

\newcommand{\add}[1]{#1} %{{\color{blue}#1}}

\newcommand{\commentout}[1]{}
\newcommand{\showArxiv}[1]{\ifthenelse{\NOT\isundefined{\ARXIV}}{#1}{}}
\newcommand{\showConfn}[1]{\ifthenelse{\NOT\isundefined{\CONFN}}{#1}{}}

\title{%
  On Correctness of Automatic Differentiation
  \\
  for Non-Differentiable Functions
}

\author{%
  Wonyeol Lee${}^\dagger$\qquad
  Hangyeol Yu${}^\dagger$\qquad
  Xavier Rival${}^\ddagger$\qquad
  Hongseok Yang${}^\dagger$\\
  ${}^\dagger$School of Computer Science, KAIST, South Korea\\
  ${}^\ddagger$INRIA Paris, D\'{e}partement d'Informatique of ENS, and CNRS/PSL University, France\\
  \hspace{-1.3em}%
  \add{%
    \texttt{\{wonyeol.lee.cs,\;yhk1344\}@gmail.com}\quad %% \texttt{yhk1344@kaist.ac.kr}\\
    \texttt{rival@di.ens.fr}\quad
    \texttt{hongseok.yang@kaist.ac.kr}%
  }\\
}

\begin{document}
%% Specify version: either ARXIV or CONFN
\newcommand{\ARXIV}{}

\maketitle
%auto-ignore

\begin{abstract}
  %% Previous version by HSY.
  %%
  %% Gradient computation or estimation lies at the core of many machine learning algorithms,
  %% and it is well-supported by popular autodiff systems, such as TensorFlow, PyTorch, and Jax.
  %% These algorithms and systems are derived for tasks involving only differentiable functions,
  %% but in practice, they are applied to models using non-differentiable functions.
  %% For instance, neural networks using ReLU as nonlinearity
  %% define non-differentiable functions in general,
  %% but the gradients of the losses of those functions are commonly computed
  %% using autodiff systems in practice.
  %% This gives rise to a natural question:
  %% are gradient-based algorithms and gradient-computing autodiff systems correct
  %% in any formal sense when they applied to non-differentiable models?
  %% In this paper, we provide a positive answer to this question
  %% and its generalised version about higher-order derivatives.
  %% In doing so, we point out a problem with an often-used informal argument
  %% that ReLU or other non-differentiable functions in deep learning do not cause an issue
  %% because their non-differentiabilities happen very rarely, i.e., they form a measure-zero set.
  %%
  Differentiation lies at the core of many machine-learning algorithms,
  and is well-supported by popular autodiff systems, such as TensorFlow and PyTorch.  % Jax.
  Originally, these %autodiff 
  systems have been developed to compute derivatives of
  differentiable functions, but in practice, they are commonly applied to 
  functions with non-differentiabilities.
  For instance, neural networks using ReLU 
  % as nonlinearity
  define non-differentiable functions in general,
  but the gradients of losses involving those functions are computed
  using autodiff systems in practice.
  This status quo raises a natural question:
  are autodiff systems correct
  in any formal sense when they are applied to such non-differentiable functions?
  In this paper, we provide a positive answer to this question.
  % and also to a similar question about higher-order derivatives.
  Using counterexamples, we first point out flaws in often-used informal arguments, such as:
  non-differentiabilities arising in deep learning 
        %(e.g., from ReLU) 
  do not cause any issues
  because %they occur rarely (i.e., 
        they form a measure-zero set.
  We then investigate a class of functions, called PAP functions,
  that includes nearly all (possibly non-differentiable) functions in deep learning nowadays.
  For these PAP functions, we propose a new type of derivatives, called intensional derivatives,
  and prove that these derivatives always exist and coincide with standard derivatives for almost 
  all inputs. We also show that these intensional derivatives are what most autodiff systems compute 
  or try to compute essentially. In this way, we formally establish the correctness of autodiff systems
  applied to non-differentiable functions.
%  We believe this is one of the first works
%  that provide such correctness of autodiff systems,
%  especially about higher-order derivatives.
%
  %% \wl{The sqrt issue in TensorFlow and PyTorch might be a minor point,
  %%   which does not need to be mentioned in the abstract?}
\end{abstract}

%% Main texts
%auto-ignore

\section{Introduction}
%
% 
% Plan -- General story
%
% \begin{enumerate}
% \item Gradient. Derivative. Driving force in modern machine learning research. Well-supported by autodiff systems, such as TensorFlow, PyTorch, Jax.
% \item But in practice, used outside of the domain of their common theoretical justification. Non-differentiable models. ReLU. Other operator. Why OK? Informal argument not enough. Previous formal studies. Interesting, but not general enough yet: no support for higher-order derivative, and no justification of gradient estimator used in the training of deep generative models. Understandable because interesting new math based on Clarke subderivative or conservative set-valued field etc, so that standard results from analysis are not directly applicable.
% \item In the paper, we propose a new justification. Justification of AD. Higher-order derivatives as well (Perhaps the ones computed by AutoGrad, Jax and Dice). Also more importantly justification of gradient estimator. Summary of results. 
% \item Key insight. Intensional view of functions. Constructed from good functions. Chain rule and higher-order derivative on this intensional representation.
% \end{enumerate}
% 

Automatic differentiation or autodiff is one of the key technologies behind the dramatic progress of deep learning in recent years \cite{RumelhartHW86,LecunBH15,Schmidhuber15-DL,BaydinPRS17}. It refers to the idea of developing and using a generic tool that can differentiate any function expressed as a program in a general-purpose programming language \cite{GriewankW08,PearlmutterS08}. Effective autodiff systems have been developed for popular programming languages
\cite{
  % Tensorflow, Pytorch
  Tensorflow16,Pytorch19,
  % Python
  Theano10,Jax18a,Tangent18,Chainer19,Autograd15,
  % Julia, F#/C#, C++, Fortran/C
  Juliadiff16,Diffsharp16,Adolc12,Tapenade13%
}.
They have enabled the development of sophisticated models and algorithms in machine learning that, in particular, involve deep neural networks~\cite{GoodfellowBC16}.

This paper is concerned with one seeming contradiction of these autodiff systems: the systems have originally been developed to compute derivatives of differentiable functions, but in practice, they are commonly applied to 
functions with non-differentiabilities. For instance, neural networks using ReLU % as nonlinearity
define non-differentiable functions in general,
but the derivatives of losses involving those functions are computed
using autodiff systems in practice.
This status quo raises 
a natural question:
are autodiff systems correct
in any formal sense when % they are 
applied to such non-differentiable functions?

A common reaction to the question is: non-differentiabilities arising in 
deep learning (e.g., from ReLU) do not cause any issues because they occur rarely 
(i.e., they form a Lebesgue-measure-zero set). In the paper, we first show that this reaction needs to be carefully re-examined 
at least. Using counterexamples, we point out flaws in three often-used arguments derived from this reaction. We then present our answer. It is also positive, but based on a class of functions that satisfy
a condition called \emph{piecewise analyticity under analytic partition} (in short, PAP). These PAP functions 
include nearly all (possibly non-differentiable) functions in deep learning nowadays.
For these PAP functions, we propose a new type of derivatives, called \emph{intensional derivatives},
and prove that these derivatives always exist and coincide with standard derivatives for almost 
all inputs. These intensional derivatives behave almost as well as, and sometimes even better than, usual derivatives for differentiable functions. For instance, they always satisfy a chain rule even if functions are non-differentiable. Using these properties of intensional derivatives, we show that the intensional derivatives are what most autodiff systems compute or try to compute essentially. In this way, we formally establish the correctness of autodiff systems that compute derivatives of non-differentiable functions.

We use $(a,b)$, $(a,b]$, $[a,b)$, and $[a,b]$ to denote intervals in $\bR$,
and $\langle a_1, \ldots, a_n \rangle$ to denote tuples.
For $n \in (\bZ_{>0} \cup \{\infty\})$, $[n]$ means the set $\{1,2,\ldots,n\}$. We call Lebesgue measure simply by measure.
The detailed statements and missing proofs of our results can be found in 
the appendix.

%auto-ignore

\section{Challenges}
\label{sec:issue}

As mentioned in the introduction, practitioners frequently apply autodiff systems to functions with non-differentiabilities, and justify these out-of-scope use cases with plausible yet heuristic arguments. In this section, we analyse these arguments. We go through three claims that are often used in the arguments implicitly, and show that although looking innocent at the outset, the claims have serious flaws; they are wrong, and we provide counterexamples.

Recall a notion of correctness
%% \footnote{\add{%
%%     There can be multiple other notions of correctness for such autodiff systems.
%%     See \S\ref{sec:conclusion} for discussion on them.%
%% }}
for an autodiff system covering non-differentiable functions~\cite{KakadeL18,BolteP20a,GriewankW08}:
% 
% 
%\xr{This definition is implicitely parameterised by the languge in which
%  programs are written (this is related to the discussion of the supported
%  program constructions, that should appear somewhere)} 
%
%\hsy{I have been aware of this. The purpose of this definition is to give 
%  a semi-formal overview on what goes on, so I feel that it is ok not 
%  to spell out a programming language.}
%  
\begin{definition}[Correctness of Autodiff]
\label{defn:correctness-autodiff}
        We say that an autodiff system is {\bf correct} if the following condition holds: for every measurable function $f : \cX \to \bR^m$ defined on an open $\cX \subseteq \bR^n$ and implemented by a program,
%\footnote{This definition is parameterised by a programming language, such as an idealised language in \S\ref{sec:autodiff}, and Python and other real-world languages.} 
if $f$ is differentiable almost everywhere (i.e., the set of inputs making $f$ non-differentiable is contained in a measure-zero set), then for almost every $x \in \cX$, the autodiff system applied to (the program of) $f$ and the input $x$ computes the derivative of $f$ at $x$.
\end{definition}
The definition permits a non-differentiable function as an input to an autodiff system, as long as its non-differentiability occurs rarely (i.e., at a measure-zero subset of the input domain). For such a function, it may be impossible to compute the correct derivative for all inputs,
simply because the derivative does not exist for inputs where the function is non-differentiable.
Thus, the definition just requires that the system should compute the correct derivative for \emph{most} inputs instead (i.e., for a subset of the input domain
whose complement inside the domain is contained in a measure-zero set).
%% that has the same measure as the domain itself).

Proving the correctness of an autodiff system is surprisingly difficult. Nearly every autodiff system is based on a chain rule for computing the derivative of function composition, but when the component functions are non-differentiable, designing a correct version of the rule is challenging. To help the reader see this challenge, we analyse three plausible yet flawed claims about the derivative of function composition, which are sometimes used implicitly in heuristic justifications of autodiff systems. 

Let $f: \cX \to \cY$ and $g : \cY \to \bR^l$ be measurable functions
defined over open sets $\cX \subseteq \bR^n$ and $\cY \subseteq \bR^m$.
%% [0,1] \to [0,1]
Here is the first claim:
\begin{description}
        \item[Claim 1] If $f$ and $g$ are differentiable almost everywhere and continuous, then $g \circ f$ should be differentiable almost everywhere. 
\end{description}
A rationale for the claim goes as follows. In order for $g \circ f$ to be non-differentiable at $x_0$, the function $f$ has to be non-differentiable at $x_0$, or it should map $x_0$ to a non-differentiable input to $g$ and be able to vary enough in a neighbourhood of $x_0$. The claim says that such an $x_0$ is rare (from the perspective of measure). Of course, the first case that $f$ is non-differentiable at $x_0$ occurs rarely by assumption. The second case seems to happen rarely as well, because the non-differentiable inputs to $g$ are rare and $f$ is continuous: because of continuity, if $f$ maps many $x_0$'s (i.e., all the $x_0$ in some non-measure-zero set) to those rare non-differentiable inputs of $g$, it should behave as a constant function in the neighbourhoods of  most of those $x_0$'s.

\newcommand{\WIDTH}{.15\textwidth}%{5.25em}
\newcommand{\HEIGHT}{5.0em}
\begin{figure}
  \centering
  %% \begin{subfigure}{\WIDTH}
  %%   \centering
  %%   \includegraphics[width=\textwidth]{{fig-response/plot_1}.pdf} \quad %,height=\HEIGHT
  %%   \caption{...}
  %% \end{subfigure}
  \includegraphics[width=\WIDTH]{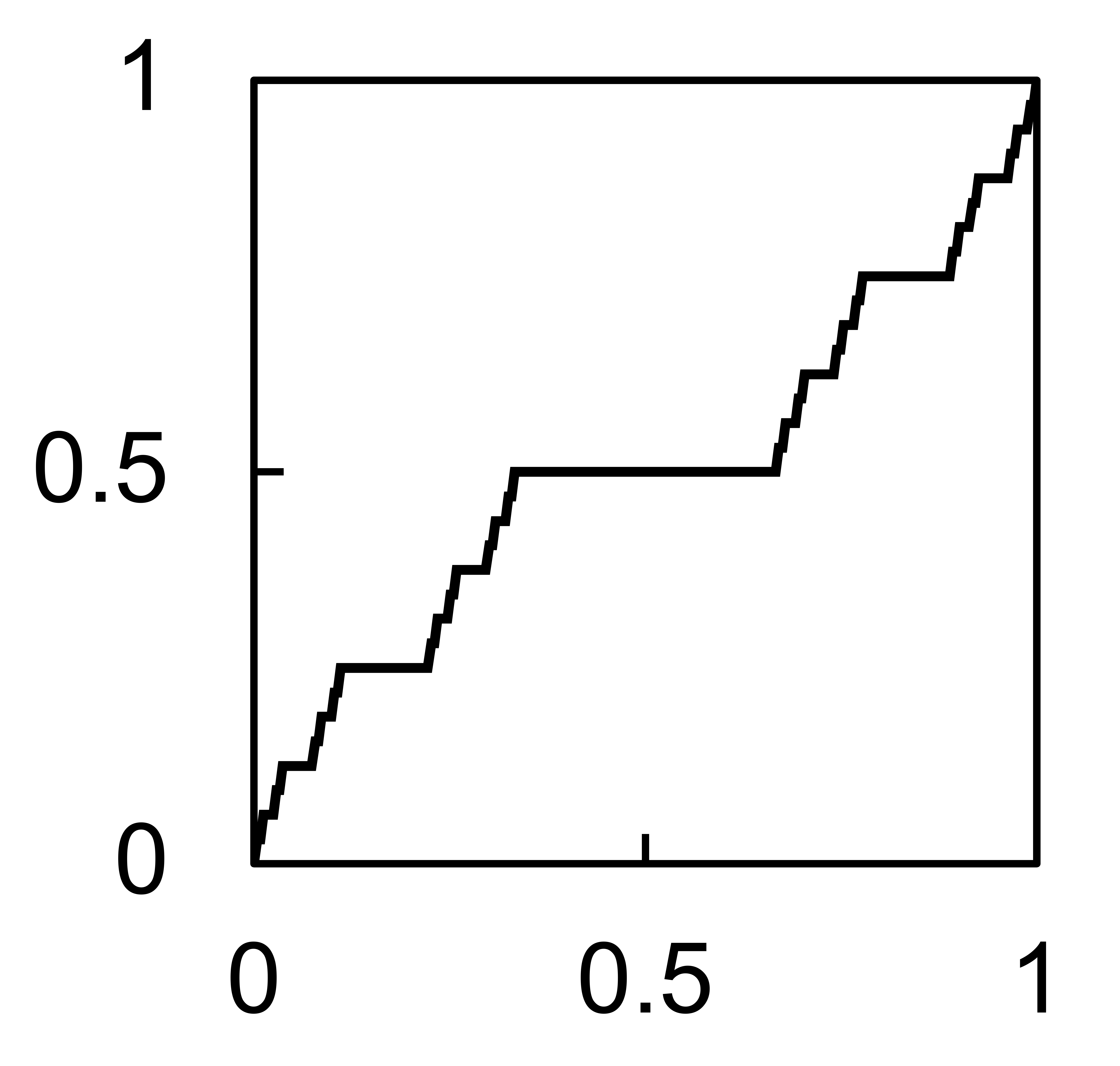} \qquad
  \includegraphics[width=\WIDTH]{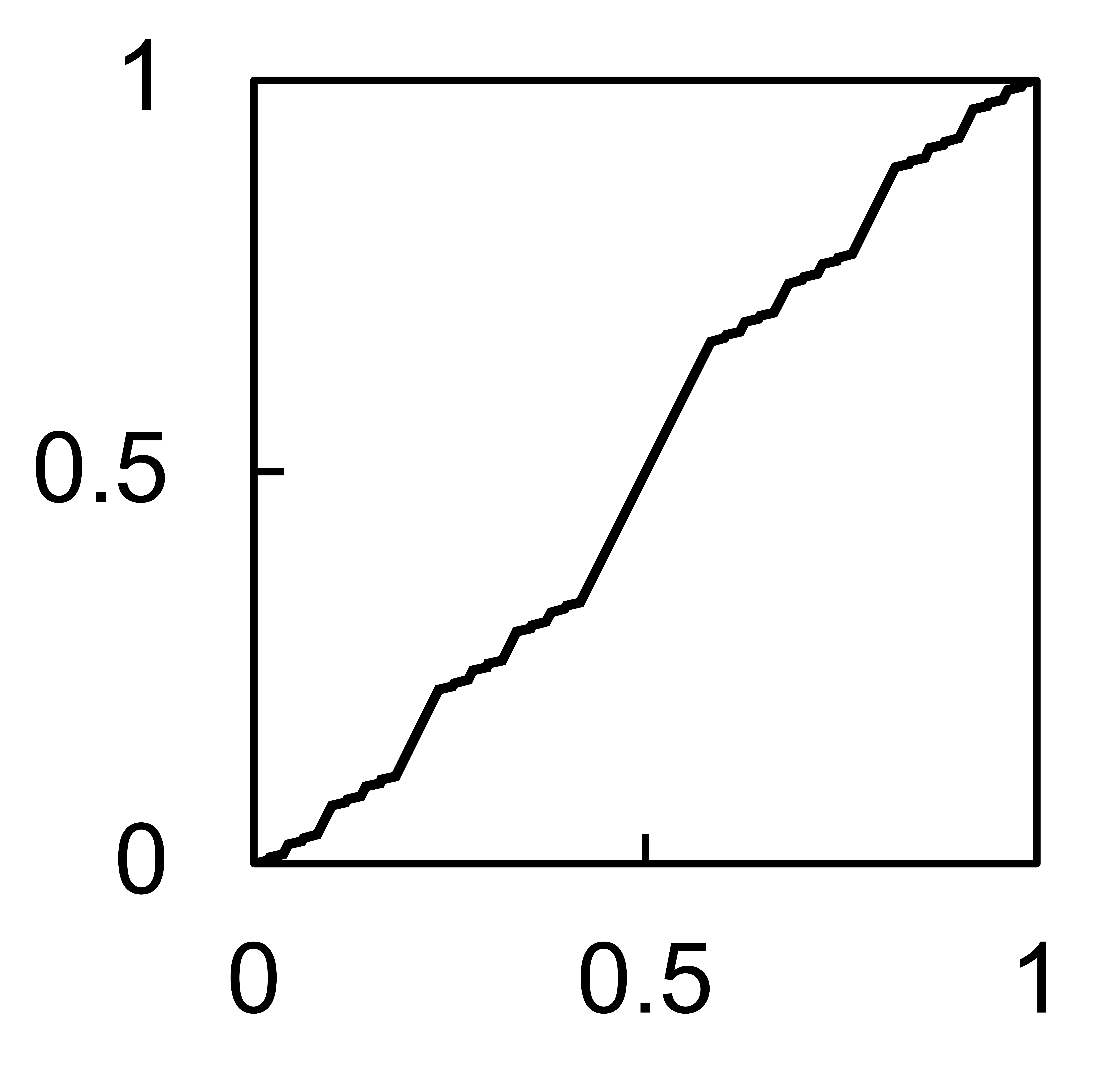} \qquad
  \includegraphics[width=\WIDTH]{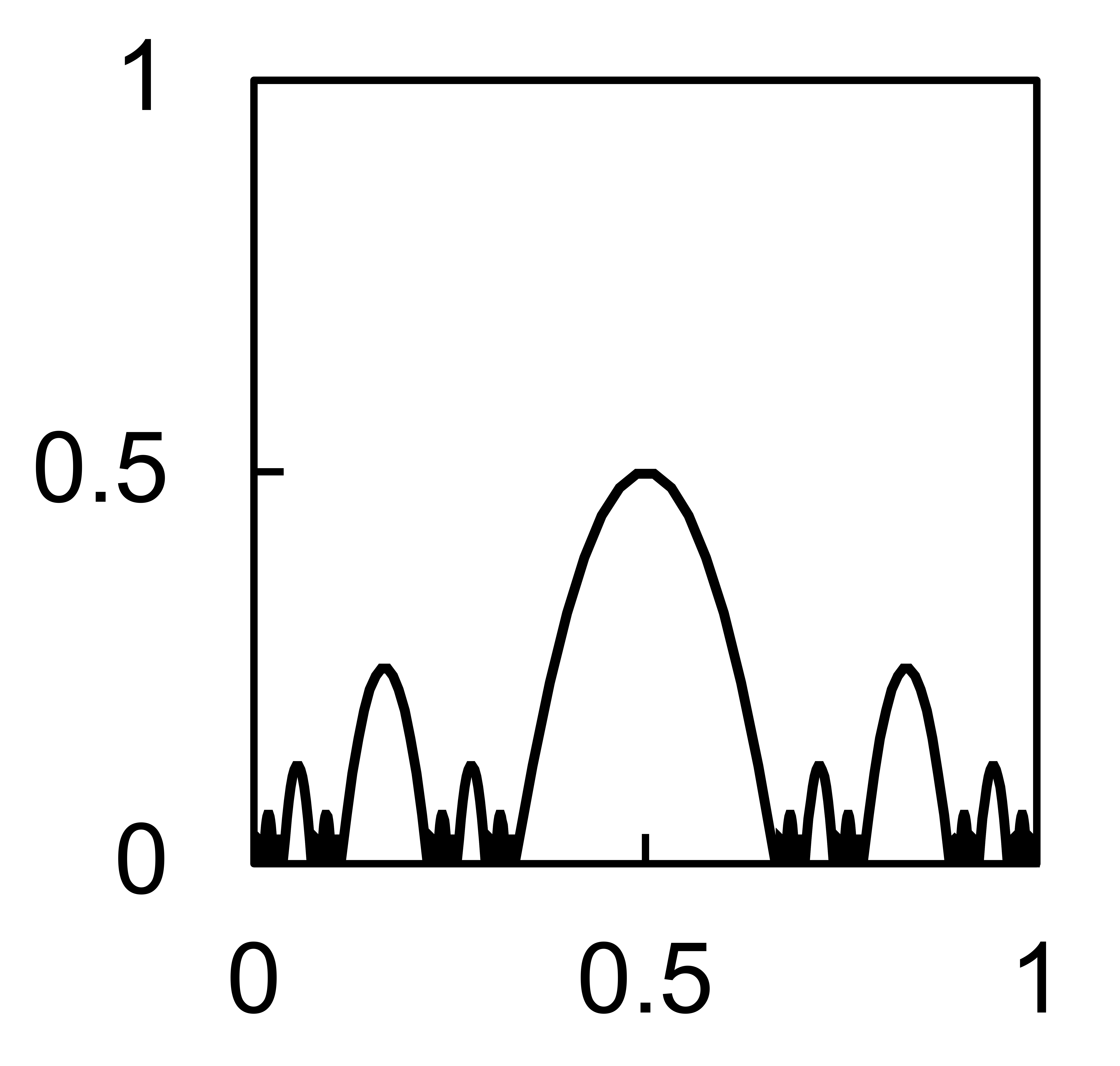} \qquad
  \caption{\add{The graphs of $\phi_1$, $f$, and $g$ that are described in the proof of \cref{prop:claim1-debunk}.}}
  \label{fig:graphs}
  %\vspace{-40pt}%
\end{figure}

The rationale has a flaw, and the claim is false. The inputs $x_0$ falling into the second case are not necessarily rare. Although $f$ is continuous, 
it is possible that $f$ maps many $x_0$'s to some of those rare non-differentiable inputs of $g$ without acting as a constant function in a neighbourhood of each of those $x_0$'s. %\footnote{The situation does not change even when $f$ is Lipschitz continuous.} 
The precise result is summarised in the following proposition:
%
%\xr{The notation of open intervals is the same as the one usually seen for
%  pairs (as 2-tuples), which is a bit distracting.}
%\wl{I also had this in mind. Any other good notation?}
%\xr{maybe $\langle a,b \rangle$ or $]a,b[$}
%\hsy{I am implementing Xavier's suggestion.}
%
\begin{proposition}
  \label{prop:claim1-debunk}
  There exist functions $f: {(0,1)} \to {(0,1)}$ and $g: {(0,1)} \to {[0,1]}$ such that
  $f$ and~$g$ are differentiable almost everywhere
  and continuous, %% (and $f$ is even Lipschitz continuous),
  but $g \circ f$ fails to be almost-everywhere differentiable.
  %(that is,
  %%% there exists a measurable subset $A \subseteq (0,1)$
  %%% such that the measure of $A$ is greater than $0$ and
  %      for some measurable subset $A \subseteq {(0,1)}$ with non-zero measure,
  %$(g \circ f)'(x)$ is undefined for all $x \in A$).
\end{proposition}
\begin{myproof}
  Before our proof, we review a generalised version of the Cantor set and Cantor function~\cite{Cantor84,Darst72} defined over the open interval ${(0,1)}$: %% \cite{Dence79}
  the $\lambda$-Cantor set $C_\lambda$ and the $\lambda$-Cantor function $\phi_\lambda$
  for $\lambda \in {(0,1]}$, which reduce to the original Cantor set and function when $\lambda=1$. The 
  $C_\lambda$ and $\phi_\lambda$ will serve as building blocks of counterexamples presented in this paper.
%
% \xr{The construction below seems to make $0$ and $1$ elements of $C_\lambda$.}
% \wl{I made the exclusion more explicit.}
%
  The set $C_\lambda \subset {(0,1)}$ consists of all the real numbers not removed during the following procedure: %% except $0$ and $1$ that are
  at step $k=0$, start with a closed interval $[0,1] \subset \bR$;
  at each step $k > 0$, remove every open interval of length $\lambda/3^k$
  which locates exactly at the middle of each remaining closed interval;
  and after all these steps, remove $0$ and~$1$.
  Note that what remains after the step $k$ are $2^k$ closed intervals.
  The function $\phi_\lambda : {(0,1)} \to {(0,1)}$ is defined as follows,
  based on the construction of $C_\lambda$.
  At each step $k > 0$, define $\phi_\lambda$
  over the $i$-th open interval to be removed
  as $\phi_\lambda(x) = (2i-1)/2^k$ ($i \in [2^{k-1}]$).
  After all these steps,
        $\phi_\lambda$ is defined over some dense open subset $S$ of $[0,1]$ (in fact, $S = (0,1) \setminus C_1$).
  Since $\phi_\lambda : S \to [0,1]$ is uniformly continuous
  with $S$ being dense in $[0,1]$,
  it has the unique continuous extension $\phi_\lambda : [0,1] \to [0,1]$
  \cite[Exercise 4.13]{Rudin76}.
  The $\lambda$-Cantor function refers to its restricted version
        $\phi_\lambda : {(0,1)} \to {(0,1)}$.
%
% \xr{it seems not obvious that the limit at the points of $C_\lambda$ is well 
%   defined so the extension is not so trivial to me.}
% \wl{Yes, it is not that trivial. I added more explanations with reference.}
%

  Returning to the proof,
%
% \xr{I think that inverse of $F$ is not defined over the whole ${(0,1)}$
%   open interval, but just on $C_{1/2}$, but that we need to extend it 
%   (it is constant over any interval $A \subseteq {(0,1)}\, \setminus C_{1/2}$).}
% \wl{No, $F$ is bijective, and not constant on $A$, due to the term $x/2$.}
%
  let $f$ be the inverse of the homeomorphism $F \,{:}\, {(0,1)} \,{\to}\, {(0,1)}$
  defined by $F(x) \,{=}\, \frac{1}{2}(\phi_1(x) + x)$ \cite[Example 2.3.1]{BoyarskyG12}.
  It is known that $f(C_{1/2})\,{=}\,C_1$ \cite[Example 2.3.2]{BoyarskyG12}.
  Construct $g : {(0,1)} \to [0,1]$ based on the construction of $C_1$ as follows,
  similarly to \cite[Example 8.18]{GelbaumO03}:
  at each step $k > 0$, define $g$ over the $i$-th open interval
  $(l_i, r_i)$ to be removed, as
  %% WL: g(x) = sqrt{...} is not necessary!
  %%\[g(x) = {1 - \Bigl(\frac{x-(r_i+l_i)/2}{(r_i-l_i)/2}\Bigr)^2}, \]
  $g(y) = 2^{-k}\cdot [1 - (y-(r_i+l_i)/2)^2/((r_i-l_i)/2)^2]$
  %%(i.e., draw a parabola with height $1$ and width $r_i-l_i$ on $(l_i, r_i)$),
  ($i \in [2^{k-1}]$);
  and define $g$ over $C_1$ as $g(y)=0$.
  \add{See \cref{fig:graphs} for the graphs of $\phi_1$, $f$, and $g$ constructed so far.}
  %% Clearly, $f$, $g$, and $g \circ f$ are continuous.
  Clearly, $f$ is continuous.
  Also, $g$ is continuous,
  since the height of the parabolas defined at the $k$-th step of $g$'s construction
  converges to $0$ as $k \to \infty$, and $g(C_1) = \{0\}$
  (see Appendix~\ref{sec:proof-issue} for the details).
  Hence, $g \circ f$ is continuous.
  Note that $f$ is even Lipschitz continuous.
  To prove this, observe that $f$ can be constructed similarly to $\phi_{1/2}$,
  and repeat the proof of the Lipschitz continuity of %% essentially 
  $\phi_{1/2}$~\cite{Darst72}. % [\S 5]

  We now show that $f$ and $g$ are differentiable almost everywhere, but $g \circ f$ is not.
  First, %% consider~$f$.
  %% WL: unnecessary anymore. =====
  %% Since $F$ is a homeomorphism, $f$ is continuous and bijective and thus monotone.
  %% Hence, $f$ is differentiable almost everywhere by the monotone differentiation theorem.
  %% =========
  since $f$ is Lipschitz continuous,
  it is differentiable almost everywhere by Rademacher's theorem~\cite[Theorem 2.2.4]{Tao11}.
  %any Lipschitz continuous function from $\bR^n$ to $\bR$
  %is differentiable almost everywhere.
  Next, %% consider $g$.
  since $g$ is differentiable on ${(0,1)}\, \setminus C_1$ by its construction,
  it is differentiable almost everywhere (as $C_1$ has measure $0$).
  Lastly, %% consider $g \circ f$.
  $g \circ f$ is non-differentiable on $C_{1/2}$, which has measure $1/2$, due to that:
  the parabolas defined at the $k$-th step of $g$'s construction
  get sharper as $k \to \infty$; $g(C_{1})=\{0\}$;
  and $f$ is a homeomorphism with $f(C_{1/2})=C_1$
  (see Appendix~\ref{sec:proof-issue} for the details).
  %% WL: wrong example and proof. ========
  %% we claim that $g \circ f$ is non-differentiable on $A = C_{1/2}$
  %% which has measure $1/2$.
  %% For the claim, it suffices to show that for any $x \in A$ and $\delta \in {(0, x)}$,
  %% there is $x' \in (x-\delta,x+\delta) \setminus \{x\}$
  %% such that $|(g \circ f)(x') - (g \circ f)(x)|=1$.
  %% Its proof is in Appendix~\ref{sec:proof-issue}, and uses the following:
  %% $f$ is strictly increasing with $f(C_{1/2})=C_1$;
  %% $C_1$ is a perfect set with no interior;
  %% and $g(C_1) = \{0\}$ and
  %% $g(I) = {(0,1]}$ for all open intervals $I$
  %% removed in the construction of~$C_1$.
  %% =========
  %% %% that $g \circ f$ is non-differentiable on $C_{1/2}$.
  %% %, and make a few comments on this proof.
  \qed
\end{myproof}
%The proposition suggests that we should look for a condition stronger than almost-everywhere differentiability, if we want to prove the correctness of an autodiff system compositionally; the correctness follows from properties of elementary functions, such as $f$ and $g$, without referring to any properties of compound functions such as $g \circ f$. We will describe such a condition later in the paper. 

The second claim is about a chain rule. 
\begin{description}
\item[Claim 2] If $f$, $g$, and $g \circ f$ are differentiable almost everywhere and continuous, then the standard chain rule for $g \circ f$ should hold almost everywhere. In particular,
  %% if $f$ and $g$ are functions on ${(0,1)}$,
  if $f,g : {(0,1)} \to {(0,1)}$,
  %% then $(g \circ f)'(x_0)$ and $g'(f(x_0)) \cdot f'(x_0)$ should be the same for almost all $x_0 \in {(0,1)}$.
  then $(g \circ f)'(x_0) = g'(f(x_0)) \cdot f'(x_0)$
  for almost all $x_0 \in {(0,1)}$.
\end{description}
Note that all of $f$, $g$, and $g \circ f$ in the claim are assumed to be differentiable almost everywhere. The claim comes from heuristic reasoning that if we just avoid those rare non-differentiable inputs of $g \circ f$, we should be able to use the standard result for differentiation, including the chain rule.

The second claim is also wrong. The flaw in the heuristic reasoning from above is that the almost-everywhere differentiability of $f$, $g$, and $g \circ f$ does not stop $g'(f(x_0))$ from being undefined for many $x_0$'s in ${(0,1)}$. This is related to the flaw in the justification for the first claim that we explained. The next proposition and its proof provide a concrete example for this phenomenon:
\begin{proposition}
  \label{prop:claim2-debunk}
  There exist functions $f, g: {(0,1)} \to {(0,1)}$ such that 
  $f$, $g$, and $g \circ f$ are differentiable almost everywhere
  and continuous,  %% (and even Lipschitz continuous),
  but it is not that %% but it is not the case that
  $g'(f(x))$ is defined for almost all $x \in {(0,1)}$.
   %(that is,
   %%% there exists a measurable $A \subseteq {(0,1)}$ such that the measure of $A$ is greater than $0$ and
   %     for some measurable subset $A \subseteq {(0,1)}$ with non-zero measure,
   %$g'(f(x))$ is undefined for all $x \in A$).
\end{proposition}
\begin{myproof}
  Let $f(x) = 1/2$ and $g(y) = {\rm ReLU}(y-1/2)+1/2$.
  %% or $g(y)=|y-1/2|+1/2, or g(y) = \sqrt{y}.
  Then, $(g \circ f)(x) = 1/2$.
  Certainly, $f$, $g$, and $g \circ f$ are differentiable almost everywhere
  and Lipschitz continuous.
  But, $g$ is not differentiable at $f(x) = 1/2$ for all $x \in (0,1)$.
  So, it is not that $g'(f(x))$ is defined for almost all $x \in (0,1)$.
  \qed
\end{myproof}

The third claim is a natural reaction to the failure of the second claim. It implements the strategy of making the chain rule in the second claim more permissive such that the counter argument of \cref{prop:claim2-debunk} no longer applies. The claim expresses a weaker version of the rule that allows one to set the derivatives of $f$ and $g$ to arbitrary values wherever $f$ and $g$ are not differentiable.
\begin{description}
        \item[Claim 3]
          %% Let $f,g$ be functions on $[0,1]$. Assume that $f$, $g$, and $g \circ f$ are all differentiable almost everywhere and continuous. Then,
                Let $f,g : {(0,1)} \to {(0,1)}$.
          If $f$, $g$, and $g \circ f$ are differentiable almost everywhere and continuous, then
                there should exist $\mathit{df}, \mathit{dg} : {(0,1)} \to \bR$ such that
          %%       the following hold: %% for almost all $x_0 \in {(0,1)}$ and $y_0 \in {(0,1)}$:
          %% %% \[
          %% %% \mathit{df}(x_0) = f'(x_0),\quad
          %% %% \mathit{dg}(y_0) = g'(y_0),\quad
          %% %% \text{and}\quad
          %% %% (g \circ f)'(x_0) = \mathit{dg}(f(x_0)) \cdot \mathit{df}(x_0).
          %% %% \]
          %% \begin{align*}
          %%   \begin{array}{cl}
          %%           (g \circ f)'(x_0) = \mathit{dg}(f(x_0)) \cdot \mathit{df}(x_0)
          %%           &\text{for almost all $x_0 \in {(0,1)}$\,;\;\;and}
          %%     \\[0.2em]
          %%           \mathit{df}(x_0) = f'(x_0) \;\;\text{and}\;\;
          %%           \mathit{dg}(y_0) = g'(y_0)
          %%           &\text{for almost all $x_0, y_0 \in {(0,1)}$.}
          %%   \end{array}
          %% \end{align*}
                $\mathit{df}(x_0) = f'(x_0)$, 
                $\mathit{dg}(y_0) = g'(y_0)$, and
                $(g \circ f)'(x_0) = \mathit{dg}(f(x_0)) \cdot \mathit{df}(x_0)$
                for almost all $x_0, y_0 \in {(0,1)}$.
\end{description}
The functions $\mathit{df}$ and $\mathit{dg}$ in the claim are the extensions of $f'$ and $g'$ that set $\mathit{df}(x)$ and $\mathit{dg}(y)$ to arbitrary values whenever $f'(x)$ and $g'(y)$ are undefined. The chain rule in the claim is phrased in terms of these extensions $\mathit{df}$ and $\mathit{dg}$, so that it does not suffer from the problem pointed out in \cref{prop:claim2-debunk}. However, this new rule is still flawed as shown in the next proposition:
\begin{proposition}
  \label{prop:claim3-debunk}
  There exist functions $f, g : {(0,1)} \to {(0,1)}$ such that
  $f$, $g$, and $g \circ f$ are differentiable almost everywhere
  and continuous, %% (and $f$ is even Lipschitz continuous),
  but for some measurable subset $A \subseteq {(0,1)}$ with non-zero measure, they satisfy the 
  following property: $f'(x) = 0$ and $(g \circ f)'(x) \neq 0$ for all $x \in A$.
\end{proposition}
\begin{myproof}
  %\hsy{Revise the proof so that it becomes reasonably readable.}
  Consider the function $f$ in the proof of \cref{prop:claim1-debunk}.
  Let $g$ be the $1$-Cantor function $\phi_1$.
  Then, $g \circ f$ is the $(1/2)$-Cantor function $\phi_{1/2}$.
  %% \cref{prop:claim1-debunk} already showed that
  We already showed
  $f$ is differentiable almost everywhere and even Lipschitz continuous.
  Since $g$ and $g \circ f$ are monotone on ${(0,1)}$,
  they are differentiable almost everywhere
  by the monotone differentiation theorem~\cite[Theorem 1.6.25]{Tao11};
  %any monotone function from $\bR$ to $\bR$
  %is differentiable almost everywhere
  and they are clearly continuous.
  We now show there exists $A \subseteq C_{1/2}$ with the desired properties.
  %% For this, examine $f'$ and $(g \circ f)'$ over $C_{1/2}$.
  %% It turns out \cite[Example 2.3.2]{BoyarskyG12}\cite{Darst72} that
  %% for almost all $x \in C_{1/2}$,
  %% \begin{align}
  %%   \label{eq:claim3-debunk-eg}
  %%   f'(x) = 0 \quad\text{and}\quad (g \circ f)'(x) = 2.
  %% \end{align}
  Since $C_{1/2}$ has measure $1/2$, it suffices to prove
  $f'(x) = 0$ and $(g \circ f)'(x) = 2$ for almost all $x \in C_{1/2}$.
  The claim indeed holds due to the following:
  $f$ and $g \circ f$ are Lipschitz, so absolutely continuous; %% Lipschitz continuous, thereby
  $f'(x)=2$ and $(g \circ f)'(x)=0$ for all $x \notin C_{1/2}$;
  $f'(x) \geq 0$ and $(g \circ f)'(x) \leq 2$ for all $x \in C_{1/2}$ whenever these derivatives exist;
  and $C_{1/2}$ has measure $1/2$.
  For the details, see~\cite[Example 2.3.2]{BoyarskyG12} and~\cite{Darst72}.
  %% Let $A = \{x \in C_{1/2} \mid \text{\eqref{eq:claim3-debunk-eg} holds for $x$}\}$.
  %% Then, $A$ satisfies the desired properties
  %% since both $C_{1/2}$ and $A$ have measure $1/2$.
  %% Hence, there exists a measure-zero set $B \subset C_{1/2}$ such that
  %% %% \eqref{eq:claim3-debunk-eg} holds for all $x \in C_{1/2} \setminus B$.
  %% $f'(x) = 0$ and $(g \circ f)'(x) = 2$ for all $x \in C_{1/2} \setminus B$.
  %% Letting $A = C_{1/2} \setminus B$ completes the proof.
  \qed
  %%Note that $f$ and $g \circ f$ are Lipschitz continuous (but $g$ is not).
\end{myproof}
The proposition implies the third claim is doomed. The claim says 
that $\mathit{df}(x_0) = f'(x_0)$ 
and 
$(g \circ f)'(x_0) = \mathit{dg}(f(x_0))\cdot \mathit{df}(x_0)$
for almost all $x_0 \in A$.
But both equalities cannot hold simultaneously:
%% By the first equality and \cref{prop:claim3-debunk}, 
if they do, by \cref{prop:claim3-debunk},
%% \begin{align*}
%%   (g \circ f)'(x_0)
%%   = \mathit{dg}(f(x_0))\cdot \mathit{df}(x_0)
%%   = \mathit{dg}(f(x_0))\cdot f'(x_0)
%%   = \mathit{dg}(f(x_0))\cdot 0 = 0,
%% \end{align*}
$(g \circ f)'(x_0)
  = \mathit{dg}(f(x_0))\cdot \mathit{df}(x_0)
  = \mathit{dg}(f(x_0))\cdot f'(x_0)
  = \mathit{dg}(f(x_0))\cdot 0 = 0,$
%% but $(g \circ f)'(x_0) \neq 0$ by the same proposition, implying the failure of the second equality.
but the same proposition also entails $(g \circ f)'(x_0) \neq 0$, leading to a contradiction.

A lesson from these flawed claims is that although the notion of correctness in \cref{defn:correctness-autodiff} only refers to almost-everywhere differentiability, we need a condition stronger than it, which behaves better in handling function composition and gives rise to a chain rule. We describe such a condition next.
%, related derivatives, and their chain rule next.

%auto-ignore

\section{PAP Function and Intensional Derivative} %% Piecewise Analytic
\label{sec:theory}

Our justification of autodiff systems relies on two key concepts: \emph{piecewise analyticity under analytic partition}, and \emph{intensional derivative}. The first is a (strictly) stronger property about functions than almost-everywhere differentiability, and yet it is satisfied by practically all programs targeted at by existing autodiff systems, as we will show in \S\ref{sec:autodiff}. Functions with this new property, called PAP functions, have an unusual type of derivatives, called intensional derivatives, which form the second concept. Intensional derivatives of PAP functions are defined everywhere and satisfy a chain rule, while still agreeing with standard derivatives for almost all inputs. In fact, the PAP functions have not just first-order but also all higher-order intensional derivatives.
%That is, they are infinitely differentiable with respect to these new derivatives. 
In \S\ref{sec:autodiff}, we will show that most autodiff systems compute intensional derivatives when applied to functions with non-differentiabilities. 

%% \wl{I think we should restrict all domains to be a open subset of $\bR^n$,
%%   not an arbitrary subset of $\bR^n$, to make proofs works.
%%   to properly handle ${\rm sqrt}: \bR_{\geq 0} \to \bR$ later, we may have to
%%   extend its domain to $\bR$ and set its function value and derivative value on $\bR_{<0}$ to $0$.}

%% For $n \in (\bZ_{>0} \cup \{\infty\})$, let $[n] = \{1,2,\ldots,n\}$.
To expand the overview of the two concepts just given, we need a notion of piecewise representation:
\begin{definition}[Piecewise Representation]
  \label{def:intens-repr}
  Let $\cX \subseteq \bR^n$ and $\cY \subseteq \bR^m$. 
        %be arbitrary sets. 
  A set of pairs $\gamma = \{\langle A^i,f^i \rangle\}_{i \in [I]}$ is a {\bf piecewise representation} of a function from $\cX$ to $\cY$ or simply {\bf representation} if and only if (i) $I \in \bZ_{>0} \cup \{\infty\}$; (ii) $\{A^i\}_{i \in [I]}$ is a partition of $\cX$; and (iii) each $f^i : \cX^i \to \cY$ is a function on an open domain $\cX^i \subseteq \bR^n$ with $A^i \subseteq \cX^i$.
% ($i \in [I]$). %% for all $i \in [I]$.
  The {\bf evaluation} of a representation $\gamma = \{\langle A^i,f^i \rangle\}_{i \in [I]}$ is the map $\eval{\gamma} : \cX \to \cY$ defined by:
        $\eval{\gamma}(x) \,{=}\, f^i(x)$ for all $i \in [I]$ and $x \in A^i$.
% \[
%          \eval{\gamma}(x) = f^i(x) \quad\text{for all $i \in [I]$ and $x \in A^i$}.
%  \]
  %% For a function $f : \cX \to \cY$, we let $\irepr{f} = \{\gamma \mid \eval{\gamma} = f\}$, the set of all representations of $f$.
\end{definition}
A %piecewise 
representation $\gamma = \{\langle A^i,f^i \rangle\}_{i \in [I]}$ describes a function $f$ in terms of a finite or countable number of component functions $f^i$ and their scopes $A^i$ of application. For each input $x$, the value of $f$ at $x$ is determined by some component $f^i$, chosen based on which piece of the input partition $\{A^i\}_{i \in [I]}$, the $x$ belongs to. Our notation for the described $f$ is $\eval{\gamma}$. Note that the domain $\cX^i$ of each $f^i$ has to be open, and it may be larger than $A^i$, the set of inputs where $f^i$ is used. As a result, a function can be described by an infinite number of representations. For instance, ReLU has at least two representations:
$\{\langle \bR_{<0},x\longmapsto 0 \rangle, \langle\bR_{\geq 0},x\longmapsto x\rangle\}$ and 
$\{\langle\bR_{<0},x\longmapsto 0\rangle, \langle\{0\},x\longmapsto 2x\rangle, \langle\bR_{> 0},x\longmapsto x\rangle\}$.

Having a piecewise representation of a function, instead of just the function, has several advantages. One of them is that for many properties of functions, we can derive their piecewise variants using those representations. We apply this advantage to analyticity. Recall that a real-valued function is analytic if and only if it is infinitely differentiable and is equal to its Taylor expansion. For $\bR^m$-valued function $g$, the analyticity means the coordinate-wise analyticity: for all $j \in [m]$, the composition $\pi_j \circ g$ of the $j$-th projection $\pi_j$ and $g$ 
is analytic. Analytic functions are favoured by autodiff systems, because they are infinitely differentiable and become zero only at a measure-zero set of inputs (unless they are zero everywhere or their domains are disconnected).
Let $\cX \subseteq \bR^n$ and $\cY \subseteq \bR^m$ be any sets.
%be arbitrary sets.
\begin{definition}[Analytic Partition]
  \label{def:analytic-partition}
  A set $A \subseteq \bR^n$ is {\bf analytic} if and only if for some $J, L \in \bZ_{>0}$, there are analytic functions $g^+_{j} : \cX^+_j \to \bR$ and $g^-_{l} : \cX^-_{l} \to \bR$
  over open domains $\cX^+_j, \cX^-_l \subseteq \bR^n$ ($j \in [J]$, $l \in [L]$) such that 
  $A = \{ x \in \bR^n \mid
  (\bigwedge_{j \in [J]} x \in \cX^+_j \land g^+_j(x) > 0)
  \land
  (\bigwedge_{l \in [L]} x \in \cX^-_l \land g^-_l(x) \leq 0)\}$.
%  \[
%  x \in A
%  \iff
%  \bigg(\bigwedge_{j \in [J]} (x \in \cX^+_j) \land (g^+_j(x) > 0)\bigg)
%  \land
%  \bigg(\bigwedge_{l \in [L]} (x \in \cX^-_l) \land (g^-_l(x) \leq 0)\bigg).
%  \]
  A partition $\{A^i\}_{i \in [I]}$ of $\cX$
  is {\bf analytic} if and only if
  all $A^i$ are analytic. %% regions inside $\cX$.
\end{definition}
\begin{definition}[PAP Representation]
  A representation
  $\gamma = \{\langle A^i,f^i\rangle\}_{i \in [I]}$ from $\cX$ to $\cY$ is {\bf piecewise analytic under analytic partition}
  (in short, {\bf PAP}) if and only if $\{A^i\}_{i \in [I]}$ is an analytic partition of $\cX$ and $f^i$ is analytic over its domain $\cX^i$ for all $i \in [I]$. 

\end{definition}
\begin{definition}[PAP Function]
  A function $f : \cX \to \cY$ is {\bf piecewise analytic under analytic partition}
        (in short, {\bf PAP}) if $f = \eval{\gamma}$ for some PAP representation $\gamma$.
\end{definition}
The definitions identify PAP representations and PAP functions as those built by the two-step process: we first split the input domain such that boundaries of the split regions are expressed by the zero sets of analytic functions, and next choose an appropriate analytic function for each piece of the split. Note the use of analytic functions in both steps. Thus, just like the standard analyticity, the PAP property implies almost-everywhere differentiability (\cref{prop:pap-ae-diff}),
%% although these two are not the same. 
but not vice versa (\cref{prop:cont-ae-diff-not-pap}).
\begin{proposition}[\cite{ZhouGKRYW19}]
  \label{prop:pap-ae-diff}
  All PAP functions are differentiable almost everywhere.
\end{proposition}
\begin{myproof}
  The proof extends %% is almost the same as
  the one for a similar result in \cite{ZhouGKRYW19}. %% [Theorem 1]
  The key idea is to use the fact that
  the zero set of a non-constant analytic function over a connected open domain
  has measure zero \cite{Mityagin15}.
  To prove the proposition,
  we show that
  for each PAP function $f$,
  there exist  countably many non-constant analytic functions $\{g_j\}_j$ over connected open domains
  such that if $f$ is non-differentiable at $x \in \cX$, then $x$ belongs to the zero set of some $g_j$.
  %% A~complete proof appears in Appendix~\ref{sec:proof-prop-pap-ae-diff}.
  For the details, see Appendix~\ref{sec:proof-prop-pap-ae-diff}.
  %% the appendix (\cref{prop:deriv-pap-func}).
  \qed
\end{myproof}
\begin{proposition}
  \label{prop:cont-ae-diff-not-pap}
  There is a continuous almost-everywhere differentiable yet non-PAP function.
\end{proposition}
\begin{myproof}
  We have two sufficient conditions for a function $f$ to be non-PAP:
  (i)
  the $k$-th order derivative of $f$ is undefined on a set of positive measure
  for some $k \geq 1$ (\cref{prop:intr-deriv-coincidence}); and
  (ii)
  the $k$-th order derivative of $f$ is undefined on an uncountable set for some $k \geq 1$,
  and $f$ is defined on a subset of $\bR$ (Appendix~\ref{sec:proof-rest}).
  %% (\cref{prop:pap-ndiff-countable} in the appendix).
  %
  The following functions satisfy (i) with $k=2$: 
        the $\lambda$-Cantor function for all $\lambda \in (0,1)$,
        $f$ in the proof of \cref{prop:claim1-debunk}, %%,prop:claim3-debunk},
        and Volterra's function $V : (0,1) \to \bR$
  \cite[Example 8.35]{GelbaumO03}. %% \cite[Exercise 5:5.5]{BrucknerBT97}. %% \cite{Volterra81}
  The following functions satisfy (ii) with $k=1$:
  the $1$-Cantor function and
  $g$ in the proof of \cref{prop:claim1-debunk}.
  All these functions are known (or already proven above)
  to be continuous and almost-everywhere differentiable.
  % (and even everywhere differentiable in case of Volterra's function).
  Hence, all of them are desired functions.
  %%For more details, refer to Appendix~\ref{sec:proof-prop-cont-ae-diff-not-pap}.
  %%
  %% WL: we no longer below argument for that \phi_\lambda (\lambd in \langle 0,1 \rangle) is not PAP.
  %%     the above argument is much simpler. =====
  %% but for $\lambda \in \langle 0,1\rangle$,
  %% we need to use the result~\cite{Dence79} which describes
  %% a sufficient condition for $x$ to be a non-differentiable input of $\phi_\lambda$,
  %% since we do not directly know whether such inputs are uncountably many.
  %% ===================
  \qed
\end{myproof}

We now define intensional derivatives of PAP representations and PAP functions.
%show that they have all the nice properties mentioned in the beginning of this section.
%% Let $\cX \subseteq \bR^n$ and $\cY \subseteq \bR^m$ be open subsets,
Let $\gamma = \{\langle A^i,f^i \rangle \}_{i \in [I]}$ be a PAP representation from $\cX$ to $\cY$ for some $\cX \subseteq \bR^n$ and $\cY \subseteq \bR^m$.
\begin{definition}[Intensional Derivative of PAP Representation]
  \label{defn: intr-derv-repr}
  %% Let $\gamma = \{\langle A^i,f^i\rangle \}_{i \in [I]}$ be a PAP representation from $\cX$ to $\cY$.
  An {\bf intensional derivative} of $\gamma$ is the set $D\gamma = \{\langle A^i, Df^i\rangle \}_{i \in [I]}$, where $Df^i$ is the standard derivative of $f^i$ viewed as a function from the domain $\cX^i$ of $f^i$ to $\bR^{m \times n}$.
\end{definition}
\begin{proposition}
  \label{prop:intr-derv-repr-wd}
  The intensional derivative $D\gamma$ is a PAP representation from $\cX$ to $\bR^{m \times n}$.
\end{proposition}
\begin{myproof}
  Nearly all the requirements for $D\gamma$ to be a PAP representation directly follow from the fact that $\gamma$ is PAP. The only exception is the analyticity of $Df^i$. There we use the fact that the operation of taking a (standard) partial derivative of a function preserves analyticity
  \cite[Proposition 2.2.3]{KrantzP02}.
  \qed
\end{myproof}
\begin{definition}[First-Order Intensional Derivative]
  \label{def:intr-deriv-fst}
  Let $f : \cX \to \cY$ be a PAP function.
  Define $\ideriv{f}$ to be the following set of functions:
  $\ideriv{f} = \{ \eval{D\gamma} \mid \text{$\gamma$ is a PAP representation of $f$}\}$.
%  \begin{align*}
%    \ideriv{f} &=
%    \{ \eval{D\gamma} \mid
%          \gamma \in \irepr{f}\ \text{and $\gamma$ is PAP}\}.
%  \end{align*}
  Each $\mathit{df} \in \ideriv{f}$ is called
  a {\bf (first-order) intensional derivative} of $f$.
\end{definition}
%{\parskip=0pt
By \cref{prop:intr-derv-repr-wd}, only PAP functions live in $\ideriv{f}$. Thus, we can also take intensional derivatives of functions in $\ideriv{f}$. We push this observation further and define higher-order intensional derivatives:
%}
\begin{definition}[Higher-Order Intensional Derivative]
  \label{def:intr-deriv-hgh}
  Let $f : \cX \to \cY$ be a PAP function.
  For each $k \in \bZ_{\geq 0}$, inductively define the set $\iderivk{k}f$ of functions by:
    $\iderivk{0}f = \{f\}$, and 
    $\iderivk{k}f =
          \{ \mathit{df}^{k} \in \ideriv (\mathit{df}^{k-1}) \mid
          \mathit{df}^{k-1} \in \iderivk{k-1}f\}\,$ for $k \geq 1$.
%  \begin{align*}
%    \iderivk{0}f &= \{f\},
%    &
%    \iderivk{k}f &=
%          \{ \mathit{df}^{k} \in \ideriv (\mathit{df}^{k-1}) \mid
%          \mathit{df}^{k-1} \in \iderivk{k-1}f\}
%    \;\;\text{for $k \geq 1$}.
%  \end{align*}
        Each $\mathit{df}^k \in \iderivk{k}f$ is called a {\bf $k$-th order intensional derivative} of $f$.
\end{definition}

A function $f : \cX \to \cY$ may have zero, one, or more than one intensional derivatives. Having at least one intensional derivative corresponds to $f$ being differentiable in the standard sense. The next propositions show that every PAP function is infinitely differentiable in the new sense (\cref{prop:intr-deriv-pap}), and that these standard and new notions of differentiability coincide if we ignore a measure-zero set of inputs (\cref{prop:intr-deriv-coincidence}).
Let $f : \cX \to \cY$ be a PAP function for some $\cX \subseteq \bR^n$ and $\cY \subseteq \bR^m$.
\begin{proposition}
  \label{prop:intr-deriv-pap}
  %% For all PAP functions $f : \cX \to \cY$ and
  For all $k \in \bZ_{\geq 0}$, the set $\iderivk{k}f$ of the $k$-th order intensional derivatives is not empty. Furthermore, its elements are again PAP functions from $\cX$ to $\bR^{m \times n^k}$.
\end{proposition}
%% \showArxiv{
%% %% ========== ARXIV ==========
\begin{myproof}
  \add{%
  The proof is by induction on $k$. First, for the case $k = 0$, $\iderivk{k}f$ is a singleton set and its unique element $f$ is PAP. Hence, the proposition holds. Next, consider the case $k > 0$. By induction hypothesis, $\iderivk{k-1}f$ is not empty and consists of PAP functions only. Thus, the set
  $\{\gamma \mid \gamma$ is a PAP representation of $g$ for some $g \in \iderivk{k-1}f\}$
is not empty. This and \cref{prop:intr-derv-repr-wd} imply that $\iderivk{k}f$ (obtained by applying $\eval{D-}$ on the above set) is nonempty and contains only PAP functions.
\qed%
  }
\end{myproof}
%% %% ========== ARXIV ==========
%% }
\begin{proposition}%%[Standard Derivative vs.~Intensional Derivative]
  \label{prop:intr-deriv-coincidence}
  %% Let $f : \cX \to \cY$ be a PAP function
        For all $k \in \bZ_{\geq 0}$, $f$ has the $k$-th order standard derivative almost everywhere,
        and this derivative agrees with any $k$-th order intensional derivative $\mathit{df}^k \in \iderivk{k}f$ almost everywhere.
\end{proposition}
{\parskip=-2pt
In the proposition, we view the $k$-th order standard derivative of $f$ as a function of type $\cX \to \bR^{m \times n^k}$. For instance, when $k=1$, the derivative maps an input to the Jacobian matrix of $f$.
}
\begin{myproof}[\cref{prop:intr-deriv-coincidence}]
  %%\hsy{Fill in the proof.}
  The first claim is proven similarly to \cref{prop:pap-ae-diff},
  except that we additionally use the following:
  an analytic function is infinitely differentiable.
  As in \cref{prop:pap-ae-diff}, we prove a stronger statement:
  there exist countably many non-constant analytic functions $\{g_j\}_j$ over connected open domains such that
  for all $k$, if the $k$-th order standard derivative of $f$ is not defined at $x \in \cX$, %%some input,
  then $x$ is in the zero set of some $g_j$. %% the input
  %% Its complete statement and proof appear in Appendix~\ref{sec:proof-prop-pap-ae-diff}.
  %% the appendix (\cref{prop:deriv-pap-func}).
  Next, consider the second claim.
  Its current form is not strong enough to enable inductive proofs.
  We instead prove a stronger statement by induction on~$k$:
        for each $\mathit{df}^k \in \iderivk{k}f$,
  there exist countably many non-constant analytic functions $\{h_l\}_l$ over connected open domains such that  
        the $k$-th order standard derivative of $f$ is well-defined, and agrees with $\mathit{df}^k$,
  at all those inputs not in the zero sets of $\{h_l\}_l$.
  %% For the details, see Appendix~\ref{sec:proof-prop-intr-deriv-coincidence}.
  %% the appendix (\cref{prop:intr-deriv-coincidence-full}).
  For the details, see Appendices~\ref{sec:proof-prop-pap-ae-diff}
  and~\ref{sec:proof-prop-intr-deriv-coincidence}.
  \qed
\end{myproof}

Intensional derivatives behave better than standard derivatives. First, the intensional derivatives of a PAP function $f$ are total functions, i.e., functions defined for all inputs (\cref{prop:intr-deriv-pap}). Contrast this with the fact that the standard derivative of $f$ is a partial function in general. The intensional derivatives can be understood as extensions of this standard derivative of $f$ to those problematic inputs that make $f$ non-differentiable (\cref{prop:intr-deriv-coincidence}). 
The totality simplifies the reasoning about intensional derivatives. 
Second, the intensional derivatives satisfy a chain rule
for all PAP functions (\cref{prop:intr-deriv-chain-rule}).
Let $f : \cX \to \cY$ and $g : \cY \to \bR^l$ be PAP functions
for some $\cX \subseteq \bR^n$ and $\cY \subseteq \bR^m$.
\begin{proposition}
  \label{prop:fn-comp-PAP}
  $g \circ f$ is a PAP function.
\end{proposition}
\begin{myproof}
   Since $f$ and $g$ are PAP, they have PAP representations
   $\gamma_f = \{\langle A^i,f^i \rangle \}_{i \in [I]}$ and
   $\gamma_g = \{\langle B^j, g^j\rangle \}_{j \in [J]}$.
   Define their composition as follows: %% $\gamma_g \circ \gamma_f$
   $\gamma_g \circ \gamma_f
     = \{\langle C^{\langle i,j\rangle}, g^j \circ f^i \rangle \}_{\langle i,j\rangle \in [I] \times [J]}$
   where $C^{\langle i,j\rangle } = \{x \in \cX \mid x \in A^i \land f^i(x) \in B^j\}$.
%   \begin{align*}
%      \gamma_g \circ \gamma_f
%      &= \{\langle C^{\langle i,j\rangle}, g^j \circ f^i \rangle \}_{\langle i,j\rangle \in [I] \times [J]},
%      &
%      C^{\langle i,j\rangle }
%      &= \{x \in \cX \mid x \in A^i \land f^i(x) \in B^j\}.
%   \end{align*}
   Then, $\gamma_g \circ \gamma_f$ is a representation 
   of $g \circ f$. Also, it is PAP
   %% mainly because the operation of composing functions preserves analyticity
   as the composition of analytic functions is analytic
   \cite[Proposition 2.2.8]{KrantzP02}.
   Thus, $g \circ f$ is PAP. %% as claimed.
   \qed
\end{myproof}
\begin{proposition}[Chain Rule for Intensional Derivatives]
  \label{prop:intr-deriv-chain-rule}
        Let $\mathit{df} : \cX \to \bR^{m \times n}$ and $\mathit{dg} : \cY \to \bR^{l\times m}$ be intensional derivatives of $f$ and $g$
        (i.e., $\mathit{df} \in \ideriv f$ and $\mathit{dg} \in \ideriv g$).
  Let $h = g \circ f$.
  Then, the following function $\mathit{dh} : \cX \to \bR^{l \times n}$ 
  is an intensional derivative of $h$ (i.e., $\mathit{dh} \in \ideriv{h}$):
  $\mathit{dh}(x)
    = \mathit{dg}(f(x)) \cdot \mathit{df}(x)$ for all $x \in \cX$,
%  \begin{align*}
%    \mathit{dh}(x)
%    &= \mathit{dg}(f(x)) \cdot \mathit{df}(x)
%    \quad\text{for all $x \in \cX$},
%  \end{align*}
        where $\mathit{dg}(f(x)) \cdot \mathit{df}(x)$ is the multiplication of matrices $\mathit{dg}(f(x))$ and $\mathit{df}(x)$.
\end{proposition}
\begin{myproof}
  By the definition of intensional derivative,
  $\mathit{df} = \eval{D\gamma_f}$ and $\mathit{dg} = \eval{D\gamma_g}$
  for some PAP representations $\gamma_f = \{\langle A^i,f^i\rangle \}_{i \in [I]}$
  and $\gamma_g = \{\langle B^j, g^j\rangle \}_{j \in [J]}$ of $f$ and $g$.
  Let $\gamma_g \circ \gamma_f$ be the composed representation
  defined in the proof of \cref{prop:fn-comp-PAP}.
  Then, $\gamma_g \circ \gamma_f$ is a PAP representation of $g \circ f$.
  So, $\eval{D(\gamma_g \circ \gamma_f)}$ is an intensional derivative of $g \circ f$.
  %% i.e., $\eval{D(\gamma_g \circ \gamma_f)} \in  \ideriv{(g \circ f)}$.
  Also, for all $x \in \cX$, if we let $\langle i,j\rangle \in [I] \times [J]$
  with $x \in A^i$ and $f^i(x) \in B^j$, then
  $\eval{D(\gamma_g \circ \gamma_f)}(x)
    = 
    D(g^j \circ f^i)(x)
    = 
    D(g^j)(f^i(x)) \cdot D(f^i)(x)
    =
    \eval{D\gamma_g}(f(x)) \cdot \eval{D\gamma_f}(x)$.
%  \begin{align*}
%    \eval{D(\gamma_g \circ \gamma_f)}(x)
%    & = 
%    D(g^j \circ f^i)(x)
%          && \text{($\because$ definition of intensional derivative)~}
%    \\
%    & = 
%    D(g^j)(f^i(x)) \cdot D(f^i)(x)
%          && \text{($\because$ chain rule for standard derivative)~}
%    \\
%    & =
%    \eval{D\gamma_g}(f(x)) \cdot \eval{D\gamma_f}(x)
%          && \text{($\because$ definition of intensional derivative).}
%  \end{align*}
  The first and last equalities follow from the definition of intensional derivative. The second equality uses the chain rule for standard derivatives:
  the rule holds here because $f^i$ and $g^j$ are analytic in neighbourhoods of $x$ and $f^i(x)$, respectively.
  \qed
\end{myproof}
We next use these properties to show that existing autodiff systems compute intensional derivatives.

%auto-ignore

\section{Correctness of Autodiff Systems}
\label{sec:autodiff}

Consider a simple programming language that assumes real-valued input variables $x_1,\ldots,x_N$ and has the following syntax for programs:
$e 
  \,::=\,
  {c}
  \,\mid\, {x}_i
  \,\mid\, \bar{\code{f}}(e_1,\ldots,e_n)
  \,\mid\, \code{if}~(e_1>0)~e_2~e_3.$
%\begin{align*}
%  e \quad
%  &::=
%  \quad {c}
%  \,\mid\, {x}_i
%  \,\mid\, \bar{\code{f}}(e_1,\ldots,e_n)
%  \,\mid\, \code{if}~(e_1>0)~e_2~e_3
%  %%~|~ \code{let}~\code{x}=e_1 ~\code{in}~ e_2,
%\end{align*}
  
A~program $e$ in the language describes a real-valued computation.
It is a real number $c$, an input variable $x_i$,
the application of a primitive function $\bar{\code{f}}$ to arguments $e_1,\ldots,e_n$,
or a conditional expression.
In the third case, the applied function is required to be a PAP function of the right type (i.e., $\bR^n \to \bR$ in this case).
We remark that practically all primitive functions supported by autodiff systems
are indeed PAP; we are unaware of any non-PAP primitive function used in practice.
If this requirement is met, all programs $e$ mean PAP functions of type $\bR^N \to \bR$. More precisely, we interpret each program $e$ as a function $\sem{e} : \bR^N \to \bR$ inductively, as shown below: for all $v \in \bR^N$,
%% \begin{align*}
%% \sem{c}v & = c,
%% \qquad\qquad
%%  \sem{x_i}v = v_i,
%% \\
%% \sem{\code{if}~(e_1>0)~e_2~e_3}v
%%         & = (\indc{\sem{e_1}v > 0}\times \sem{e_2}v) + (\indc{\sem{e_1}v \leq 0} \times \sem{e_3}v),
%% \\
%% \sem{\code{f}(e_1,\ldots,e_n)}v 
%%         & = \bar{\code{f}}(\sem{e_1}v,\ldots,\sem{e_n}v),
%% \end{align*}
%\begin{gather*}
%  \sem{c}v = c,
%  \qquad
%  \sem{x_i}v = v_i,
%  \qquad
%  \sem{\bar{\code{f}}(e_1,\ldots,e_n)}v 
%  = {\code{f}}(\sem{e_1}v,\ldots,\sem{e_n}v),
%  \\
%  \sem{\code{if}~(e_1>0)~e_2~e_3}v
%  =
%  \text{if } (\sem{e_1}v > 0)  \text{ then } \sem{e_2}v \text{ else } \sem{e_3}v
%\end{gather*}
\vspace{1mm}\newline%
\centerline{$
\begin{array}{@{}c@{}}
  \sem{c}v = c,
  \qquad
  \sem{x_i}v = v_i,
  \qquad
  \sem{\bar{\code{f}}(e_1,\ldots,e_n)}v 
  = {\code{f}}(\sem{e_1}v,\ldots,\sem{e_n}v),
  \\[0.3em]
  \sem{\code{if}~(e_1>0)~e_2~e_3}v
  =
  \text{if } (\sem{e_1}v > 0)  \text{ then } \sem{e_2}v \text{ else } \sem{e_3}v.
\end{array}
$}\vspace{1mm}\newline%
where $v_i$ is the $i$-th component of the vector $v$,
and ${\code{f}} : \bR^n \to \bR$ is the PAP function % of type $\bR^n \to \bR$
denoted by the function symbol $\bar{\code{f}}$ in the language.
Then, the defined $\sem{e}$ is always PAP. %% (\cref{prop:semantics-pap}).
\begin{proposition}
\label{prop:semantics-pap}
For every program $e$, its denotation $\sem{e}$ is a PAP function from $\bR^N$ to $\bR$.
\end{proposition}
\showArxiv{
%% ========== ARXIV ==========
\begin{myproof}
        The proof is by induction on the structure of $e$. The cases of $e \equiv c$ and $e \equiv x_i$ follow from the fact that both constant functions and projections are PAP. For the case of $e \equiv \bar{\code{f}}(e_1,\ldots,e_n)$, we note two facts about PAP functions: the composition of two PAP functions is PAP (\cref{prop:fn-comp-PAP}); and for PAP functions $g_1,\ldots,g_n : \bR^N \to \bR$, the function $v \longmapsto \langle g_1(v),\ldots,g_n(v) \rangle$ of type $\bR^N \to \bR^n$ is PAP again,
mainly because any finite intersection of open sets is again open.
By these two facts, the claimed property of the proposition holds in this case.

The only remaining case is $e \equiv (\code{if}~(e_1>0)~e_2~e_3)$.
By induction hypothesis, all of $\sem{e_1}$, $\sem{e_2}$, and $\sem{e_3}$ are PAP.
Let $\gamma_1=\{\langle A^i,f^i \rangle\}_{i \in [I]}$, $\gamma_2=\{\langle B^j,g^j \rangle\}_{j \in [J]}$,
and $\gamma_3=\{\langle C^k,h^k \rangle\}_{k \in [K]}$ be their PAP representations,
and define their conditional composition $\ifelse{\gamma_1}{\gamma_2}{\gamma_3}$ as follows:
%% For each $i \in [I]$, define $A^i_1 = \{v \in A^i \mid f^i(v) > 0\}$ and $A^i_0 = \{v \in A^i \mid f^i(v) \leq 0\}$. Also, for all $\langle i,j,k,l\rangle \in ([I]\times [J] \times [K] \times \{0,1\})$, let
%% $D^{\langle i,j,k,l \rangle} = A^i_l \cap B^j \cap C^k$ and 
%% \[
%%         t^{\langle i,j,k,l \rangle} : \dom(f^i) \cap \dom(g^j) \cap \dom(h^k) \to \bR,
%%         \qquad
%%         t^{\langle i,j,k,l \rangle}(v) = 
%%         \begin{cases}
%%                 g^j(v) & \text{if $l=1$}
%%                 \\
%%                 h^k(v) & \text{if $l=0$}.
%%         \end{cases}
%% \]
%% \begin{align*}
%%   \ifelse{\gamma_1}{\gamma_2}{\gamma_3}
%%   &= \rlap{$\{\langle E^{\langle i,j,k,l \rangle},t^{\langle i,j,k,l \rangle} \rangle\}_{\langle i,j,k,l \rangle \in ([I]\times [J] \times [K] \times \{0,1\})},$}
%%   \\
%%   E^{\langle i,j,k,l \rangle} &= A^i_l \cap B^j \cap C^k,
%%   &
%%   t^{\langle i,j,k,l \rangle} &= \text{if } (l=1) \text{ then } g^j \text{ else } h^k
%%   \qquad\qquad
%% \end{align*}
$\ifelse{\gamma_1}{\gamma_2}{\gamma_3}
= \{\langle E^{\langle i,j,k,l \rangle},t^{\langle i,j,k,l \rangle} \rangle\}_{\langle i,j,k,l \rangle \in ([I]\times [J] \times [K] \times \{0,1\})}$,
$E^{\langle i,j,k,l \rangle} = A^i_l \cap B^j \cap C^k$, and
$t^{\langle i,j,k,l \rangle} = \text{if } (l=1) \text{ then } g^j \text{ else } h^k$,
where $A^i_1 = \{v \in A^i \mid f^i(v) > 0\}$ and $A^i_0 = \{v \in A^i \mid f^i(v) \leq 0\}$.
Then, $\{E^{\langle i,j,k,l \rangle}\}_{\langle i,j,k,l \rangle}$ is an analytic partition of $\bR^N$, every $t^{\langle i,j,k,l \rangle}$ is an analytic function, and its domain is an open set containing $E^{\langle i,j,k,l \rangle}$. Thus,
%% $\gamma = \{\langle E^{\langle i,j,k,l \rangle},t^{\langle i,j,k,l\rangle} \rangle\}_{\langle i,j,k,l \rangle}$
$\ifelse{\gamma_1}{\gamma_2}{\gamma_3}$
is a PAP representation. Furthermore, its evaluation
%% $\eval{\gamma}$
$\eval{\ifelse{\gamma_1}{\gamma_2}{\gamma_3}}$
is equal to $\sem{\code{if}~(e_1>0)~e_2~e_3}$. Hence, the proposition holds in this case.
  \qed
\end{myproof}
%% ========== ARXIV ==========
}

%\xr{some parts of the sentence seems to be missing ?}
We show that if an autodiff system for the language satisfies two requirements to be described shortly, it essentially computes an intensional derivative, even when the input function is not differentiable. 
The first requirement is that
for every primitive operation $\bar{\code{f}}$ of type $\bR^n \to \bR$,
the system should come with a function $\mytilde{D}\code{f} : \bR^n \to \bR^{1 \times n}$
that satisfies  $\mytilde{D}\code{f} \in \ideriv{\code{f}}$ and serves as a ``derivative'' of $\code{f}$.
To describe the second requirement,
we inductively define the function $\semdiff{e} : \bR^N \to \bR^{1 \times N}$,
formalising symbolic differentiation of a program $e$, as follows:
for all $v \in \bR^n$,
\vspace{1mm}\newline%
\centerline{$
  \begin{array}{@{}c@{}}
  \semdiff{c}v
  = \vec{0}_{1 \times N},
  \quad
  \semdiff{\bar{\code{f}}(e_1,\ldots,e_n)}v
  = 
  (\mytilde{D}\code{f})(\sem{e_1}v,\ldots,\sem{e_n}v)%^{\top}\!
  \cdot [ \semdiff{e_1}v; \,\ldots; \,\semdiff{e_n}v ],
  \\[0.3em]
  \semdiff{x_i}v
          \,{=}\, [\vec{0}_{(i-1) \times 1};\vec{1}_{1\times 1};\vec{0}_{(N-i)\times 1}]^\top\!\!,
  \ \;
  \semdiff{\code{if}\,(e_1{>}0)\,e_2\,e_3}v
  \,{=}\, \text{if}\,(\sem{e_1}v {>} 0)\,
      \text{then}\,{\semdiff{e_2}}v
      \text{ else}\,{\semdiff{e_3}}v.
\end{array}
$}\vspace{1mm}\newline%
%\begin{align*}
%  &
%  \mytilde{D}(c)v
%  = \vec{0}_{N},
%  \qquad
%  \mytilde{D}(\bar{\code{f}}(e_1,\ldots,e_n))v
%  = 
%  (\mytilde{D}\code{f})(\sem{e_1}v,\ldots,\sem{e_n}v) 
%        \cdot \langle \mytilde{D}(e_1)v, \ldots, \mytilde{D}(e_n)v\rangle,
%  \\
%  &
%  \mytilde{D}({x_i})v
%  = \langle\vec{0}_{i-1}, 1, \vec{0}_{N-i}\rangle,
%  \qquad
%  \mytilde{D}(\code{if}~(e_1\,{>}\,0)~e_2~e_3)v
%  = \text{if } (\sem{e_1}v \,{>}\, 0) 
%      \text{ then } {\mytilde{D}(e_2)}v 
%      \text{ else } {\mytilde{D}(e_3)}v.
%\end{align*}
Here 
$\vec{c}_{i \times j}$ denotes the $i\times j$ matrix containing only the constant $c$, and
the RHS of the second equation means the multiplication of
the $1 \times n$ matrix $(\mytilde{D}\code{f})(\sem{e_1}v,\ldots,\sem{e_n}v)$ and
the $n\times N$ matrix $[ \semdiff{e_1}v; \ldots; \semdiff{e_n}v]$
that is constructed by concatenating $n$ matrices $\semdiff{e_1}v, \ldots, \semdiff{e_n}v$ of size $1 \times N$.
%% The definition of $\semdiff{e}v$ formalises a variant of the standard forward autodiff algorithm.
The definition of $\semdiff{e}v$
computes a ``Jacobian'' matrix of $\sem{e}$ (which is a $1 \times N$ matrix in this case) in the usual way of symbolic differentiation:
run the program $e$ under the input $v$, and calculate
the ``Jacobian'' matrix of the encountered operations
using the chain rule with the ``derivatives'' $\mytilde{D}\code{f}$ of those operations $\code{f}$
\cite{BaydinPRS17,WangZDWER19,AbadiP20,HuotSV20}.
%% %% prev ver by HSY:
%% The second requirement says that
%% given a program $e$ and an input $v \in \bR^N$,
%% if the system performs forward autodiff, 
%% it should output the inner product of $v$ and the vector $\semdiff{e}v \in \bR^N$ defined below:
%% [eqn]
%% On the other hand, the requirement says, if the system performs backward autodiff and thus additionally takes $y \in \bR$ (as a degenerate vector for the real-valued output), then the system should return the gradient $a \cdot \semdiff{e}v \in \bR^N$ for the input.
%%
%% WL: references for below contents and terms. ======
%% - https://arxiv.org/pdf/1811.05031.pdf
%%   - what forward/reverse autodiff computes.
%%   - initial tangent/cotangent.
%% - https://jax.readthedocs.io/en/latest/notebooks/autodiff_cookbook.html#Jacobian-Vector-products-(JVPs,-aka-forward-mode-autodiff)
%%   - jacobian-vector/vector-jacobian product.
%% ===============
%%
Our second requirement is that
given a program $e$ and an input $v \in \bR^N$,
if the system performs forward-mode (or reverse-mode) autodiff
with a tangent vector $w \in \bR^N$ (or a cotangent vector $u \in \bR$),
it should output the Jacobian-vector product $(\semdiff{e}v) \cdot w \in \bR$
(or the vector-Jacobian product $u^{\top} \!\!\cdot (\semdiff{e}v) \in \bR^{N}$) %% 1 \times N
\cite{GriewankW08,Maclaurin16,Margossian19}.
%% WL: NOTE ON REFERENCE. ======
%% - all of the below terms appear in JAX tutotiral webpage:
%%   - https://jax.readthedocs.io/en/latest/notebooks/autodiff_cookbook.html#How-it's-made:-two-foundational-autodiff-functions
%%   - but it may not be adequate to cite a webpage. so I cite the below papers/books instead.
%% - new terms and cites:
%%   - ``tangent/cotangent vector'': Margossian19 (+ GriewankW08)
%%   - ``Jacobian-vector/vector-Jacobian product'': Macluarin16 (+ GriewankW08)
%% ==============
Intuitively, the second
requirement says that the output of an autodiff system should coincide with that of
%% this variant of the forward autodiff algorithm.
symbolic differentiation.
Note that the requirement does not fix the choice of an autodiff algorithm,
and has two separate conditions for forward-mode and reverse-mode algorithms.
%, but leaves it to an autodiff system in consideration. 
Our next results show that these requirements ensure the correctness of an autodiff system.
\begin{theorem}
  \label{thm:autodiff-correct-fst}
  %% For every program $e$, we have $\semdiff{e} \in \ideriv{\sem{e}}$.
  If $\mytilde{D}\code{f} \in \ideriv{{\code{f}}}$\,
  %% for every primitive-function symbol $\bar{\code{f}}$ in the language,
  for all primitive functions $\bar{\code{f}}$,
  then $\semdiff{e} \in \ideriv{\sem{e}}$\, for all programs $e$.
\end{theorem}
\showArxiv{
%% ========== ARXIV ==========
\begin{myproof}
        The proof is by induction on the structure of $e$. When $e \equiv c$, the trivial partition $\{\bR^N\}$ and the constant function $v\longmapsto c$ form a PAP representation of $\sem{e}$. The intensional derivative of this representation is $\{\langle \bR^N, v \longmapsto \vec{0}_{1 \times N}\rangle \}$, and its evaluation is $\semdiff{c}$, as claimed by the theorem. The other base case is $e \equiv x_i$. We use the trivial partition again with the projection function $v \longmapsto v_i$, and form a PAP representation of $\sem{e}$. The intensional derivative of this representation is $\{\langle \bR^N, v \longmapsto [ \vec{0}_{(i-1) \times 1}; \vec{1}_{1 \times 1}; \vec{0}_{(N-i) \times 1}]^\top \rangle\}$, and its evaluation is $\semdiff{x_i}$. 

        The next case is $e \equiv \bar{\code{f}}(e_1,\ldots,e_n)$. By induction hypothesis, $\semdiff{e_i}$ is an intensional derivative of  $\sem{e_i}$ for every $i \in [n]$. Let $g : \bR^N \to \bR^n$ and $\mathit{dg} : \bR^N \to \bR^{n \times N}$ be functions defined by $g(v) = \langle \sem{e_1}v,\ldots,\sem{e_n}v\rangle$ and $\mathit{dg}(v) = [ \semdiff{e_1}v;\ldots;\semdiff{e_n}v ]$ for all $v$. Then, $\mathit{dg}$ is an intensional derivative of $g$. Also, $\sem{\bar{\code{f}}(e_1,\ldots,e_n)} = {\code{f}} \circ g$. Therefore, by the chain rule for intensional derivative (\cref{prop:intr-deriv-chain-rule}), the function $v \longmapsto  (\mytilde{D}\code{f})(g(v)) \cdot \mathit{dg}(v)$ is an intensional derivative of $\sem{\bar{\code{f}}(e_1,\ldots,e_n)} = {\code{f}} \circ g$.
        Here we use the assumption that  %%a part of the first requirement that
        $\mytilde{D}\code{f}$ is an intensional derivative of $\code{f}$.
        Note that the function is equal to $\semdiff{\bar{\code{f}}(e_1,\ldots,e_n)}$. So, the theorem holds.

The last case is $e \equiv (\code{if}~(e_1>0)~e_2~e_3)$. By induction hypothesis, $\semdiff{e_i}$ is an intensional derivative of $\sem{e_i}$ for all $i \in [3]$.
Let $\gamma_1$, $\gamma_2$, and $\gamma_3$ be the PAP representations of
$\sem{e_1}$, $\sem{e_2}$, and $\sem{e_3}$ such that
%% \begin{equation}
%%   \label{thm:autodiff-correct-fst:eqn1}
%%   \eval{D\gamma_2} = \semdiff{e_2}
%%   \quad\text{and}\quad
%%   \eval{D\gamma_3} = \semdiff{e_3}.
%% \end{equation}
$\eval{D\gamma_2} = \semdiff{e_2}$ and $\eval{D\gamma_3} = \semdiff{e_3}$.
Let $\ifelse{\gamma_1}{\gamma_2}{\gamma_3}$ be
the conditionally composed representation defined in the proof of \cref{prop:semantics-pap}.
Then, it is a PAP representation of $\sem{\code{if}~(e_1>0)~e_2~e_3}$.
%% But by the equations in \eqref{thm:autodiff-correct-fst:eqn1} and the definitions of
But by the definitions of $\gamma_2$, $\gamma_3$,
$\semdiff{-}$, $\ifelse{-}{-}{-}$, and $\eval{D-}$,
we have $\eval{D\ifelse{\gamma_1}{\gamma_2}{\gamma_3}}(v)
= \semdiff{\code{if}~(e_1>0)~e_2~e_3}v$
for all $v \in \bR^N$. Hence, the theorem holds in this case as well.
%
%\xr{The following two constructions on partitions are used in several places;
%  how about defining them and instantiating them without redoing them each time:
%  (1) coarser partition that is finer than two partitions
%  (2) reverse image of a partition by a function
%  (it might save space and make the arguments clearer)
%}
%\wl{For a part of this, I newly defined $\ifelse{-}{-}{-}$.}
%\xr{Thanks, this is similar to what I meant, it is nice.}
\qed
\end{myproof}
%% ========== ARXIV ==========
}
%\begin{corollary}
%        If $D\code{f} \in \ideriv{\bar{\code{f}}}$\, for every primitive-function symbol $\code{f}$ in the language,
%        then for all programs $e$, $\mytilde{D}(e)v$ is the derivative of $\ideriv{\sem{e}}(v)$ at almost all $v \in \bR^N$.
%\end{corollary}
\begin{corollary}
        \label{cor:autodiff-correctness-fst}
        Assume that an autodiff system for the language in this section satisfies the two requirements. 
        Then, for each program $e$, there exists an intensional derivative $\mathit{df}$
        of $\sem{e}$ %% in $\ideriv{\sem{e}}$
        %% %% prev ver by HSY:
        %% such that if the system performs forward autodiff, it computes the inner product $\langle v, \mathit{df}v\rangle$
        %% for every input $v \in \bR^N$, and if the system performs backward autodiff, it computes the vector $a \cdot \mathit{df}v$
        %% when given $a \in \bR$. Furthermore, in both cases, the computed entities are the corresponding inner product
        %% and the constant-vector multiplication with the standard derivative of $\sem{e}$ for almost all inputs $v$ in $\bR^N$.
        such that if the system performs forward-mode (or reverse-mode) autodiff
        with a tangent vector $w \in \bR^N$ (or a cotangent vector $u \in \bR$),
        it computes the Jacobian-vector product $\mathit{df}(v) \cdot w \in \bR$
        (or the vector-Jacobian product $u^{\top} \!\!\cdot \mathit{df}(v) \in \bR^{N}$) %% 1 \times N
        for every input $v \in \bR^N$.
        Furthermore, the computed entity is
        the corresponding Jacobian-vector product (or vector-Jacobian product)
        with the standard derivative of $\sem{e}$ for almost all inputs $v \in \bR^N$.
\end{corollary}
\showArxiv{
%% ========== ARXIV ==========
\begin{myproof}
  Suppose that the two requirements are met, and let $e$ be a program.
  Consider the case when the system performs forward-mode autodiff.
  Let $\mathit{df} = \semdiff{e}$.
  Then, by Theorem~\ref{thm:autodiff-correct-fst},
  $\mathit{df}$ is an intensional derivative in $\ideriv{\sem{e}}$.
  Moreover, by the second requirement,
  the output of the system for $e$ with a tangent vector $w \in \bR^N$
  is $(\semdiff{e}v) \cdot w = \mathit{df}(v) \cdot w \in \bR$ for all inputs $v \in \bR^N$.
  This proves the first part of the corollary.
  The other part of the corollary follows immediately from \cref{prop:intr-deriv-coincidence}.
  The proof for the case when reverse-mode autodiff is performed is essentially the same,
  so we omit it.
  \qed
\end{myproof}
%% ========== ARXIV ==========
}

\begin{remark}[Intensional Derivatives in Practice]
{\rm%
  We briefly discuss whether the first requirement mentioned earlier
  is indeed met by autodiff systems.
  TensorFlow~\cite{Tensorflow16} and PyTorch~\cite{Pytorch19}, two popular autodiff systems,
  %% \cite{Pytorch17}
  support a wide range of primitive functions that are not differentiable for some inputs.
  A well-known example is $\codett{relu} : \bR \to \bR$ (i.e., the map $x \longmapsto \max(0,x)$).
%  which acts as the identity function on positive reals,
%  but as the constant zero function on non-negative reals.
  This function is not differentiable at $0$,
  but when these autodiff systems are applied to differentiate $\codett{relu}$ at $0$,
  they return $0$ instead of an error (i.e., $(\mytilde{D}\codett{relu})(0)=0$).
  It means that the systems compute the intensional derivative $\eval{D\gamma}$ of $\codett{relu}$
  for $\gamma = \{\langle \bR_{>0},x \longmapsto x\rangle, \langle \bR_{\leq 0},x \longmapsto 0\rangle \}$.
  Thus, they fulfil the first requirement on $\codett{relu}$.
  Note that setting the derivative at the input $0$ to 
  %   $1$ (i.e., $(\mytilde{D}\codett{relu})0 = 1$)
  any $c \in \bR$ is an acceptable option;
  it can be justified by a different PAP representation $\gamma'$ of $\codett{relu}$
  %% that takes the identity function on $\bR_{\geq 0}$ and the constant zero function on $\bR_{< 0}$.
  where 
  %$\gamma' = \{\langle \bR_{\geq 0},x \longmapsto x\rangle, \langle \bR_{< 0},x \longmapsto 0\rangle\}$
  $\gamma' = \{\langle \bR_{> 0},x \longmapsto x\rangle,  \langle\{0\}, x \longmapsto c\cdot x\rangle, \langle\bR_{< 0},x \longmapsto 0\rangle\}$.
  Another example is $\codett{reciprocal\_no\_nan}$ from TensorFlow. It means the map
  %% $f(x)=\indc{x \neq 0} \cdot (1/x) + \indc{x = 0} \cdot 0$.
  $f(x)=1/x$ if $x \neq 0$, and $0$ if $x=0$.
  Differentiating this map by TensorFlow gives the function
  %% $\mathit{df}(x) = \indc{x \neq 0} \cdot (-1/x^2) + \indc{x = 0} \cdot 0$.
  $(\mytilde{D}f)(x) = -1/x^2$ if $x \neq 0$, and $0$ if $x= 0$.
  The $\mytilde{D}f$ is an intensional derivative of $f$,
  because $\gamma = \{\langle \bR \setminus \{0\}, x \longmapsto 1/x\rangle,
  \langle \{0\},x \longmapsto 0 \rangle\}$
  is a PAP representation of $f$ and $\eval{D\gamma}$ coincides with $\mytilde{D}f$.
  Hence, TensorFlow meets the first requirement on $\codett{reciprocal\_no\_nan}$.

  However, for some functions, TensorFlow and PyTorch behave more conservatively
  than our theory of PAP functions permits. A good example is $\codett{sqrt}$.
  The domain of $\codett{sqrt}$ is $\bR_{\geq 0}$,
  but differentiating $\codett{sqrt}$ at $0$ using these autodiff systems
  returns $\texttt{+inf}$, which means $\infty$ (i.e., $(\mytilde{D}\codett{sqrt})(0)=\infty$).
  Thus, the first requirement on $\codett{sqrt}$ is violated by these systems.
  One consequence of this is that
  when the systems are applied to differentiate ${\tt {sqrt}({mult}(x,0))}$,
  %%{\tt sqrt(mult(x,0))},
  they end up with computing $\texttt{0} \texttt{*} (\texttt{+inf})$
  that evaluates to $\texttt{NaN}$ (meaning an undefined number in floating point arithmetic)
  for all $x \in \bR$. Note that ${\tt sqrt(mult(x,0))}$ is constantly zero and has zero as its derivative.
  This conservative behaviour could have been avoided if TensorFlow and PyTorch had recognised that
  $\codett{sqrt}$ is a PAP function and had used its intensional derivative instead.
  For instance,
  %% $d\codett{sqrt}(x) = \indc{x > 0}\cdot 1/(2\sqrt{x}) + \indc{x = 0} \cdot 0$
  the function $(d\codett{sqrt})(x) = 1/(2\sqrt{x})$ if $x > 0$, and $0$ if $x=0$,
  is an intensional derivative of $\codett{sqrt}$,
  since $\gamma = \{\langle \bR_{>0}, x \longmapsto \sqrt{x}\rangle,
  \langle \{0\},x \longmapsto 0 \rangle\}$
  is a PAP representation of $\codett{sqrt}$.
  If TensorFlow and PyTorch used $d\codett{sqrt}$
  as a derivative of $\codett{sqrt}$,
  %(i.e., if $\mytilde{D}\codett{sqrt} = d\codett{sqrt}$),
  they would be able to differentiate ${\tt {sqrt}({mult}(x,0))}$
  correctly for all $x \in \bR$.%
  %%
  %% \wl{Mention that actual implementations of relu, recip, and sqrt
  %%   and those of their ``derivatives'' can be unrelated.
  %%   Add the details (e.g., the table in rebuttal) to the appendix, and refer to it here.}
}
\end{remark}

%auto-ignore

\section{Related Work \add{and Discussion}}
\label{sec:conclusion}

%% \wl{Regarding autodiff literature, \cite{BaydinPRS17,AbadiP20}
%%   would give enough previous works to be cited below.
%%   Done with \cite{KakadeL18,BolteP20a,DavisDKL20}.}

%% {\bf Related Work.}
%% =====
%% autodiff for non-diff'l funcs [Griewank]
%% =====
Autodiff has a long history with a large body of literature
\cite{GriewankW08,BaydinPRS17,PearlmutterS08}. Its community has been aware of some issues with non-differentiable functions
\cite[Chapter 14]{GriewankW08}.
These issues have become ever more important,
as autodiff has been increasingly applied to a variety of non-differentiable functions,
including sophisticated linear algebra functions
\cite{KakadeL18,Maclaurin16,SeegerHDL17}. %% \wl{any other examples?}.
In this paper, we investigate the issues in a more systematic and rigorous way,
by presenting non-trivial concrete counterexamples
that illuminate subtleties of non-differentiable functions in autodiff,
%% how autodiff systems could go wrong when non-differentiable functions come into play;
and also proposing intensional derivatives, a new notion of derivatives,
that enable us to formally prove the correctness of,
and better understand the behaviour of, autodiff systems
applied to non-differentiable functions.

Clarke subdifferential \cite{Clarke90,ClarkeLSW98} and related notions have been favourite tools for studying the use of non-differentiable functions in the context of autodiff and stochastic gradient descent~\cite{KakadeL18,DavisDKL20,BolteP20a,Majewski18,GriewankWFB16}.
%One of the most well-studied approaches
%to handle non-differentiable functions in the context of autodiff (or stochastic gradient descent)
%\cite{KakadeL18,DavisDKL20,BolteP20a,Majewski18,GriewankWFB16} is based on
%Clarke subdifferential \cite{Clarke90,ClarkeLSW98} (or related notions).
%% \wl{\cite{Majewski18,GriewankWFB16} are taken from conservative field paper.
%%   I added them just because they look relevant; I do not know much about their contents.}
Let $f$ be an input program (or function) to be differentiated.
Assume that $f$ is locally Lipschitz continuous but possibly non-differentiable at some inputs.
%% ======
%% autodiff for subdifferential [Kakade]
%% ======
\citet{KakadeL18} proposed a linear-time algorithm that
correctly computes a {\em subdifferential} of $f$ at $x$ {\em almost surely}, for {\em all} inputs $x$,
when $f$ satisfies some qualification condition.
Their work, however, could not provide the correctness of the current autodiff systems,
because those systems do not use their algorithm.
%% %% ======
%% %% convergence of stochastic subdifferential method [Davis]
%% %% ======
%% \wl{I omitted detailed explanation of \cite{DavisDKL20},
%%   since it seems not that relevant to our work.
%%   You can find my writing in latex comments.}
%% \citet{DavisDKL20} pursue a similar line of research, proving that
%% every limit point of the stochastic gradient descent with subdifferentials
%% is a critical point of $f$ almost surely, if $f$ is Whitney stratifiable.
%% Although a breakthrough, the result still has a gap from the current autodiff systems
%% since what they compute is not subdifferential.
%% ======
%% conservative fields [Bolte]
%% ======
\citet{BolteP20a} presented a result closest to ours:
standard autodiff algorithms correctly compute the {\em standard derivative} of $f$ at
$x$ {\em always}, for {\em almost all} inputs $x$, if $f$ is definable~\cite{DriesM96}.
%% in some o-minimal structure (Corollary 5).
To do so, they extended Clarke subdifferential
to define a new notion of derivatives, called conservative fields,
and showed that autodiff algorithms compute conservative fields
and the fields coincide with the standard derivatives almost everywhere.
We point out three key differences between these works and ours.
%Our work differs from the aforementioned works using subdifferential-related notions.
First, they use highly advanced  mathematical tools 
(e.g., variational analysis \cite{RockafellarW98}).
Thanks to those powerful tools, they could prove more results than ours,
such as the almost-sure convergence of the stochastic gradient descent
with conservative fields~\cite{BolteP20a}, but they are less accessible than 
our work, which requires much simpler mathematics (probably at an undergraduate level)
but still gives the correctness proof of autodiff systems.
Second, subdifferentials and conservative fields are difficult
to be generalised to higher-order derivatives, because they are 
defined in terms of a {\em single} {\em set-valued} function.
In contrast, our intensional derivatives are defined in terms 
of a {\em set} of {\em standard} (i.e., singleton-valued) functions,
and thus have a natural generalisation to higher-order derivatives.
Lastly, different function classes are considered:
those works apply to locally Lipschitz continuous functions with additional properties
(e.g., definability), whereas ours applies to PAP functions.
We point out that PAP functions include many non-locally-Lipschitz functions,
such as discontinuous $x \in \bR \longmapsto \indc{x > 0}$
%% and continuous $x \in \bR_{\geq 0} \longmapsto \sqrt{x}$;
and continuous \add{$x \in \bR \longmapsto \sqrt{|x|}$};
%% and $x \in \bR \longmapsto x^{1/3}$
the opposite direction is also true (e.g., the $\lambda$-Cantor functions for $\lambda \in (0,1)$).
%We have not yet found an example that is non-PAP and locally Lipschitz and has the additional properties (e.g., definability)
%required by those works (e.g., \cite{BolteP20a}).

%% ======
%% piecewise-smooth [Zhou], piecewise-analytic [Griewank]
%% ======
The PAP property is similar to some other piecewise properties, %% is not completely new.
such as piecewise smoothness \cite{ZhouGKRYW19} which inspired PAP,
and piecewise analyticity \cite{GriewankW08}.
The PAP, however, differs from the others in a few ways:
unlike piecewise smoothness,
it allows piecewise functions to be defined on open subsets of $\bR^n$
(not just on $\bR^n$), but restricts them to be analytic (rather than smooth);
unlike piecewise analyticity,
it allows piecewise domains to be countably many (not just finitely many),
but restricts them to be analytic (rather than unrestricted).
Unlike prior works, we also give an in-depth study of the PAP property,
such as its closedness under operations (e.g., composition) %% basic operations
and its relationship to other properties (e.g., almost-everywhere differentiability).
%% relationship to intensional derivatives.
%%
%% ======
%% LF-PPL [Zhou], SPCF [Mak]
%% ======
%% The property of almost-everywhere differentiability of programs has been studied as well.
%% For instance, it is shown to hold always in
%% a first-order probabilistic programming language~\cite{ZhouGKRYW19,GriewankW08},
%% and in a higher-order probabilistic programming language with recursion~\cite{MakOPW20}.
The property of almost-everywhere differentiability has been studied as well
for programs and functions (that need not be related to programs).
E.g., it is shown to hold
for all programs in some programming languages \cite{GriewankW08,MakOPW20},
%% first-order and higher-order languages
and for all piecewise smooth functions \cite{ZhouGKRYW19}.
We prove the property %% (and its higher-order derivative counterpart)
for all PAP functions and utilise it in proving that
higher-order intensional derivatives agree with standard derivatives almost everywhere.
%% Extending our results to a more expressive language, such as the one in \cite{MakOPW20},
%% is a good future research direction.
%% Nesting autodiff systems has been also widely studied to compute higher-order derivatives
%% \cite{PearlmutterS08,SiskindP08,Pearlmutter94}.

%% \wl{Looks unimportant.}
%% Higher-order derivative with autodiff.
%% DiCE~\cite{Dice18}. Jax~\cite{Jax18a}.

\add{%
  Recently and concurrently with this work, \citet{BolteP20b} studied some concepts and results similar to ours.
  They proposed a new class of functions and a new notion of derivatives,
  called elementary selections and selection derivatives,
  which roughly correspond to our PAP functions and intensional derivatives;
  and proved properties of those new functions and derivatives,
  which roughly correspond to our
  \cref{%
    prop:intr-deriv-coincidence,%
    prop:fn-comp-PAP,%
    prop:intr-deriv-chain-rule,%
    prop:semantics-pap,%
    thm:autodiff-correct-fst%
  }.
  Although having some similarities, our work and their work have three key differences, complementing each other.
  First, their work is applicable to a strictly smaller class of functions than ours,
  as any elementary selection is PAP (and locally Lipschitz) but not vice versa.
  Second, it considers selection derivatives of first order only,
  whereas our work considers intensional derivatives of higher orders as well.
  Third, their work provides some results not in our work
  (e.g., convergence of stochastic gradient descent with selection derivatives),
  and vice versa (e.g., the results in \S\ref{sec:issue}).
  %%
  %% * correspondence
  %% - PAP functions <--> elementary selections.
  %% - (first-order) intensional derivative <--> selection derivative.
  %% - Proposition 8 (for $k=1$) <--> Proposition 3.
  %% - Proposition 9  <--> Proposition 1.
  %% - Propotision 10 <--> Proposition 4.
  %% - Proposition 11 <--> Proposition 5.
  %% - Theorem     12 <--> Theorem 2.
  %%
  %% * differences
  %% - the class of elementary selections: strict subset of PAP and locally Lipschitz functions;
  %%   so does not consider the use of analytic functions other than exp and log,
  %%   and does not allows any discontinuity (e.g., from branches).  
  %% - selection derivative: only first-order one is studied.
  %% - using tools from conservative fields, prove more about a.s. convergence of SGD following selection derivatives.
  %%   <--> using simple tools, give elementary proof of correctness of autodiff (for wider class of funcs),
  %%        but does not prove anything about SGD.
}

\add{%
  We conclude the section by discussing a few remarks on a notion of correctness for autodiff systems.
  First, there can be multiple different correctness conditions of autodiff systems
  and we do not claim that our correctness condition
  (\cref{defn:correctness-autodiff} or \cref{cor:autodiff-correctness-fst}) is ``the'' gold standard.
  Rather we are just suggesting ``a'' correctness condition
  that can serve as a reasonable (possibly minimal) requirement for existing and future autodiff systems. 
  Second, as important as the correctness of autodiff systems is
  the correctness of the applications built upon autodiff systems
  (e.g., gradient descent, variational inference, and Hamiltonian Monte Carlo),
  but the latter might not necessarily follow from the former.
  For example, consider an autodiff system that is correct in the sense of \cref{cor:autodiff-correctness-fst}
  (i.e., computes an intensional derivative),
  and a gradient descent algorithm that follows the ``derivative'', %%(i.e., an intensional derivative),
  computed by the system, of an objective function $f : \bR \to \bR$ to minimise $f$. %% \sem{e}
  Suppose that $f$ is given by $f(x) = x^2$ if $x \neq 1$, and $1$ if $x=1$; %% $e = \code{if}~(x_1=1)~1~x_1^2$  
  the autodiff system computes $\mathit{df}(x) = 2x$ if $x \neq 1$, and $0$ if $x=1$,
  as the ``derivative'' of $f$; and the gradient descent starts at $x=1$.%
  \footnote{\add{We thank an anonymous reviewer for coming up with this example.}}
  Then, the gradient descent gets stuck at $x=1$ forever since $\mathit{df}(1)=0$,
  even though $1$ is never a critical point of $f$.
  This example illustrates that the correctness of autodiff systems
  does not necessarily imply the correctness of gradient descents built upon those systems
  (where the latter correctness is given by that gradient descents should converge to Clarke critical points).%
  \footnote{\add{%
      On the other hand, we conjecture that
      for any PAP and locally Lipschitz $f : \bR^n \to \bR$,
      if a gradient descent follows an intensional derivative of $f$
      and starts at {\em randomly chosen} initial points,
      then it would be correct {\em almost surely}.
      A partial result confirming the conjecture was proven recently in~\cite{BolteP20b}, concurrently with this work.%
  }}
  Nevertheless, the two notions of correctness discussed so far,
  one for autodiff systems and the other for applications using autodiff systems,
  address two separate issues and this paper is mainly about the former notion of correctness.%
  %%
  %% \begin{itemize}[nosep, leftmargin=1em]
  %% \item 
  %%   Hamiltonian Monte Carlo (HMC) and variational inference (VI), possibly for probabilistic programming: We reiterate that PAP functions enjoy a nice property that they are analytic on each piece of domain, whose boundary is measure-zero. The property has been crucially used to design various methods of HMC and VI for non-differentiable densities and prove their correctness (e.g., [1,\,2]). This signifies the importance of studying PAP functions. Whether the correctness claims in those works would still hold if intensional derivatives are used in place of standard ones, is another open problem. We will discuss these interesting open problems in the revised version of the paper.
  %% \end{itemize}
  %%
  %% Although there remain a few open problems, we strongly believe that our work would serve as an important first step towards understanding and resolving those problems. Above all, as far as we know, this is the first work that (i) raises subtleties in the well-known chain rule when applied to almost-everywhere differentiable functions, (ii) gives concrete counterexamples illuminating those subtleties, and (iii) proves some reasonable (possibly minimal) correctness of existing autodiff systems that permits non-differentiable functions, using only elementary mathematics.
  %%
  %% {[1]} Discontinuous Hamiltonian Monte Carlo for discrete parameters and discontinuous likelihoods. Biometrika, 2020. \\
  %% {[2]} Reparameterization Gradient for Non-Differentiable Models. In NeurIPS, 2018.
}

%% \section{Conclusion}
%\wl{Removed conclusion (at least for now).}

%% TODO for next version:
%% - cite: [1] Sam Staton's paper on correctness of autodiff.
%%         [2] draft from conservative field people.
%% - mention: our intrinsic derivative is different from all other subdifferentials [2].

%% Impact and Bibliography
%auto-ignore

\section*{Broader Impact}

\add{%
This work focuses mainly on theoretical aspects of autodiff systems.
In particular, we formally prove that the systems,
though developed to handle differentiable functions,
remain correct even when applied to non-differentiable functions.
Our result justifies, at least in part, the current situation in machine learning,
in which the systems are frequently applied to non-differentiable functions
without much consideration to their correctness under such out-of-scope use cases.
Other than the justification,
this work does not present any other foreseeable societal consequence due to its theoretical nature.%
}

\begin{ack}  
  \add{%
    We thank anonymous reviewers for their insightful and constructive comments.
    Lee, Yang, and Yu were supported by the Engineering Research Center Program through the National Research Foundation of Korea (NRF) funded by the Korean Government MSIT (NRF-2018R1A5A1059921), and also by Next-Generation Information Computing Development Program through the National Research Foundation of Korea (NRF) funded by the Ministry of Science, ICT (2017M3C4A7068177).
    Rival was supported by a Facebook gift and by the European Research Council
    (ERC) under the European Union’s Horizon 2020 research and innovation
    programme (grant agreement No 825492).
    %% Rival was supported by \wl{Please add if you have any}.
  }
\end{ack}

\bibliography{refs}
\bibliographystyle{abbrvnat}
\addcontentsline{toc}{section}{\refname}

%% Appendix
\clearpage
\appendix
%auto-ignore

\showConfn{%
\newcommand{\supptitle}{%
  Supplementary Material:
  \\
  On Correctness of Automatic Differentiation
  \\
  for Non-Differentiable Functions%
}
% from neurips_2020.sty
\newcommand{\toptitlebar}{
  \hrule height 4pt
  \vskip 0.25in
  \vskip -\parskip%
}
\newcommand{\bottomtitlebar}{
  \vskip 0.29in
  \vskip -\parskip
  \hrule height 1pt
  \vskip 0.09in%
}
\vbox{%
  \hsize\textwidth
  \linewidth\hsize
  \vskip 0.1in
  \toptitlebar
  \centering{\LARGE\bf \supptitle\par}
  \bottomtitlebar
}%
}

%auto-ignore

\section{Comments on Results in \S\ref{sec:issue}}
\label{sec:proof-issue}

\subsection{Comments on the proof of \cref{prop:claim1-debunk}} %% Proposition
\label{sec:proof-issue-claim1}

First, we elaborate on our proof in \cref{prop:claim1-debunk}
that $g$ is continuous on $(0,1)$.
Since $g$ is continuous on $(0,1) \setminus C_1$ by its construction,
we only need to show that $g$ is continuous on $C_1$.
Consider any $x \in C_1$ and $\epsilon > 0$.
It suffices to show that there is $\delta \in (0, x)$ such that
\begin{align}
  \label{eq:proof-claim1-1}
  0 < |x - x'| < \delta \quad\implies\quad |g(x)-g(x')| = |g(x')| = g(x') < \epsilon.
\end{align}
Let $k >0$ be an integer with $2^{-k} < \epsilon$.
Consider the set \[S = \{ x' \in (0,1) \setminus C_1 \mid g(x') \geq  2^{-k} \}.\]
By the construction of $g$,
$S$ is the union of some finitely many closed intervals in $(0,1)$
that do not contain $x$.
(Note that each of those closed intervals is contained in
an open interval removed at some $k'(\leq k)$-th step of $g$'s construction.)
Hence, \[\delta = \min \left( \inf_{x' \in S}|x-x'|, \; x \right)\] is positive.
We now show that $\delta$ satisfies~\eqref{eq:proof-claim1-1}.
Consider any $x'$ with $0<|x-x'|<\delta$.
If $x' \in C_1$, then $g(x') = 0$.
If $x' \notin C_1$, then $x' \in (0,1) \setminus C_1$ and $x' \notin S$ by the definition of $\delta$,
and thus $g(x') < 2^{-k} < \epsilon$ by the definition of $S$ and $k$.
Hence, \eqref{eq:proof-claim1-1} holds and this completes the proof.
\qed

Second, we elaborate on our proof in \cref{prop:claim1-debunk}
that $g \circ f$ is not differentiable on $C_{1/2}$.
Consider any $x \in C_{1/2}$.
It suffices to show that for any $\delta \in (0,x)$,
there exist $x_1, x_2 \in (x-\delta, x+\delta) \setminus \{x\}$ such that
\begin{align}
  \label{eq:proof-claim1-2}
  \left| \frac{(g \circ f)(x) - (g \circ f)(x_1)}{x-x_1} \right|
  &= 0
  \qquad\text{and}\qquad
  \left| \frac{(g \circ f)(x) - (g \circ f)(x_2)}{x-x_2} \right|
  > 1.
\end{align}
Consider any $\delta \in (0,x)$.
Since $x \in C_{1/2}$ is a limit point of $C_{1/2}$,
there exists $x_1 \in (x-\delta, x+\delta) \setminus \{x\}$ with $x_1 \in C_{1/2}$.
For this $x_1$, the first equality in~\eqref{eq:proof-claim1-2} holds,
since $(g \circ f)(C_{1/2}) = g(C_1) = \{0\}$.
To find $x_2$, let $k>0$ be an integer such that
\begin{align}
  \label{eq:proof-claim1-3}
  d(k) = 
  \frac{1}{2} \cdot \frac{1}{2^k} +
  \frac{3}{4} \cdot \frac{1}{3^k}
  < \delta
  \qquad\text{and}\qquad
  \frac{3}{4} \cdot \left(\frac{2}{3}\right)^k < \frac{1}{2}.
\end{align}
We claim that there exists $x_2 \in (0,1)$ such that
\begin{align}
  \label{eq:proof-claim1-4}
  0 < |x - x_2| \leq d(k)
  \qquad\text{and}\qquad
  (g \circ f)(x_2) = 2^{-k}.
\end{align}
If the claim holds, then
\begin{align*}
  \left| \frac{(g \circ f)(x) - (g \circ f)(x_2)}{x-x_2} \right|
  &\geq {2^{-k}} / d(k)
  && \text{(by \eqref{eq:proof-claim1-4} and $(g \circ f)(x) = 0$)}
  \\
  &=
  2^{-k} \Big/
  \left( \frac{1}{2} \cdot \frac{1}{2^k} + \frac{3}{4} \cdot \frac{1}{3^k}  \right)
  && \text{(by the equality in \eqref{eq:proof-claim1-3})}
  \\
  & = 1 \Big/ \left( \frac{1}{2} + \frac{3}{4} \cdot \left(\frac{2}{3}\right)^k \right)
  \\
  & > 1 / \left( \frac{1}{2} + \frac{1}{2} \right) = 1
  && \text{(by the inequality in \eqref{eq:proof-claim1-3})},
\end{align*}
and thus the second equality in~\eqref{eq:proof-claim1-2} holds.
Hence, finding $x_2 \in (0,1)$ satisfying~\eqref{eq:proof-claim1-4} completes the proof.
We now show that such $x_2$ exists.
Consider the situation right after the $k$-th step of $C_{1/2}$'s construction is performed.
Then, the total length of the closed intervals that still remain is
\[
1 - \frac{1}{2}\cdot\left(\frac{1}{3^1} \cdot 2^0 + \frac{1}{3^2} \cdot 2^1
+ \cdots + \frac{1}{3^k} \cdot 2^{k-1} \right)
= \frac{1}{2} \cdot\left( 1 + \left(\frac{2}{3}\right)^k \right),
\]
so the length of each of those closed intervals is
$\frac{1}{2} (2^{-k} + 3^{-k})$,
since those closed intervals have the same length and there are $2^k$ such intervals.
Due to this, and by the construction of $C_{1/2}$,
there is some open interval $I$ that is removed exactly at the $k$-th step
of $C_{1/2}$'s construction and satisfies
\[{\rm dist}(x, I) \leq \frac{1}{2} \cdot \left( \frac{1}{2^{k}} + \frac{1}{3^{k}} \right).\]
Let $x_2 \in (0,1)$ be the midpoint of $I$.
By the construction of $f$ and $g$,
we have $(g \circ f)(x_2) = 2^{-k}$.
Furthermore, since the length of $I$ is $3^{-k}/2$,
we have \[|x - x_2| \leq {\rm dist}(x,I) + \frac{1}{2} \cdot {\rm len}(I)
\leq \frac{1}{2} \cdot \left( \frac{1}{2^{k}} + \frac{1}{3^{k}} \right)
+ \frac{1}{4} \cdot \frac{1}{3^{k}}
= d(k).\]
Hence, $x_2 \in (0,1)$ satisfies~\eqref{eq:proof-claim1-4},
and this concludes the proof.
\qed

%% WL: wrong example, since g is not continuous. ==========
%% Since $f$ maps $C_{1/2}$ onto $C_1$ and it is strictly increasing, we have $y = f(x) \in C_1$ and $y \in (f(x-\delta), f(x+\delta))$.  Since each element of $C_1$ is a limit point of $C_1$, there exists $y'' \in (f(x-\delta),f(x+\delta)) \setminus \{y\}$ such that $y'' \in C_1$.  Since $C_1$ has no interior, there exists a point in between $y$ and $y''$ that is not included in $C_1$.  By the construction of $g$, there exists $y'$ in between $y$ and $y''$ such that $g(y') = 1$.  Setting $x' = f^{-1}(y')$, we have $x' \in (x-\delta,x+\delta) \setminus \{x\}$ and $|(g \circ f)(x') - (g \circ f)(x)|=|g(y') - g(y)|=|1-0|=1$.
%% ===================

Next, we make a remark on non-differentiable inputs of $f$, $g$, and $g \circ f$ in the proof. One might guess that $f$ should be non-differentiable exactly on $C_{1/2}$, given that $f$ maps ${(0,1)}\, \setminus C_{1/2}$ onto ${(0,1)}\, \setminus C_1$ in a linear way and maps $C_{1/2}$ onto $C_{1}$ in a non-smooth-looking way.  Surprisingly, the guess is wrong: $f$ is in fact non-differentiable only on a measure-zero subset of $C_{1/2}$.  On the other hand, $g$ and $g \circ f$ are non-differentiable exactly on $C_1$ and $C_{1/2}$, respectively.
The proof that $g$ is non-differentiable on $C_1$
is similar to the above proof that $g \circ f$ is non-differentiable on $C_{1/2}$,
and thus we omit it.
  %% due to the following:
  %% $g$ is differentiable on ${(0,1)}\, \setminus C_1$ obviously;
  %% $g$ is non-differentiable on $C_1$ by an argument similar to that for $g \circ f$ presented above;
  %% $g \circ f$ is differentiable on ${(0,1)}\, \setminus C_{1/2}$
  %% since $f$ is differentiable on ${(0,1)}\, \setminus C_{1/2}$
  %% and $g$ is differentiable on $f({(0,1)}\, \setminus C_{1/2}) = {(0,1)}\, \setminus C_{1}$.

Finally, we connect the examples in the proof with our results in \S\ref{sec:theory}.
Both $f$ and $g$ are shown to be non-PAP (\cref{prop:cont-ae-diff-not-pap}).
Hence, our results do not guarantee that $g \circ f$ is PAP and so almost-everywhere differentiable.  In fact, $g \circ f$ is non-PAP, since $g \circ f$ is not almost-everywhere differentiable.

% \xr{PAP is not defined yet; I guess the introduction will remedy that.}
% \wl{I added the last two paragraphs (and other last paragraphs in other proofs of this section)
%      as a remark, with an intention to put it into the appendix;
%      or any other good idea?
% \hsy{I am adding all these in the appendix.}     
%

\subsection{Comments on the proof of \cref{prop:claim2-debunk}} %% Proposition 

We explain how the counterexample in the proof does not contradict to our results in \S\ref{sec:theory}. The functions $f$ and $g$ are PAP (and thus $g \circ f$ is so). Although $g'$ is undefined at $0$, we can extend it to an intensional derivative $\mathit{dg} \in \ideriv{g}$ such that $\mathit{dg}$ is defined everywhere (even at $0$) and coincides with $g'$ at all but countably many inputs. With such $\mathit{dg}$, the following version of the chain rule holds almost everywhere:
\[(g \circ f)'(x) = \mathit{dg}(f(x)) \cdot f'(x) \quad\text{ for almost all } x \in {(0,1)}.\]
This is because we have the chain rule for intensional derivatives and and these intensional derivatives and standard derivatives coincide almost everywhere (\cref{prop:intr-deriv-chain-rule,prop:intr-deriv-coincidence}).

\subsection{Comments on the proof of \cref{prop:claim3-debunk}} %% Proposition 

The functions $f$, $g$, and $g \circ f$ in the proof do not contradict to our results. Neither $f$ nor $g$ is a PAP function (\cref{prop:cont-ae-diff-not-pap}). Hence, our results do not guarantee the validity of our version of the chain rule for $g \circ f$. 

%% \xr{Suggestion: keep the "connection to our results" paragraphs and
%%     move them into the PAP section; I think they can illustrate how
%%     PAP "repairs" the issues}
%% \wl{I wish we could do this, but due to page limit, we could not add them to PAP section.}

%auto-ignore

\section{Comments on and Proofs for Results in \S\ref{sec:theory}}
%% \label{sec:proof-theory}
Let $\cX, \cX_f, \cX_g \subseteq \bR^n$ and $\cY \subseteq \bR^m$ be arbitrary sets.

\subsection{Comments on the proof of \cref{prop:cont-ae-diff-not-pap}} %% Proposition
\label{sec:proof-prop-cont-ae-diff-not-pap}

We prove the following argument used in the proof of \cref{prop:cont-ae-diff-not-pap}:
the functions listed in the proof satisfy the sufficient condition (i) or (ii)
mentioned in the proof.

First, consider (i).
To show that a function $h : \cX \to \bR^m$ satisfies (i) with $k=2$,
it suffices to show the claim that
the set \[S = \{x \in \cX \mid h' \text{ is undefined or discontinuous at } x\}\]
has positive measure.
For the $\lambda$-Cantor function $\phi_\lambda$ with $\lambda \in (0,1)$,
$S$ is a full measure subset of $C_\lambda$ due to the following:
$\phi_\lambda'(x)  = 1/(1-\lambda) \neq 0$ for almost all $x \in C_\lambda$;
$\phi_\lambda'(x) = 0$ for all $x \notin C_\lambda$;
and $C_\lambda \subset (0,1)$ has no interior.
Since $C_\lambda$ has measure $1-\lambda > 0$, the claim holds for $\phi_\lambda$.
For $f$  in the proof of \cref{prop:claim1-debunk},
$S$ is a full measure subset of $C_{1/2}$ due to similar reasons.
So the claim holds for $f$.
For Volterra's function,
$S$ is known to have positive measure~\cite[Example 8.35]{GelbaumO03}.
So the claim holds for Volterra's function.

Next, consider (ii).
Observe that the $1$-Cantor function and $g$ in the proof of \cref{prop:claim1-debunk}
are both defined on $(0,1) \subseteq \bR$
and non-differentiable exactly on $C_1$
(for the non-differentiability of $g$, see \S\ref{sec:proof-issue-claim1}).
Since $C_1$ is uncountable,
the two functions satisfy (ii) with $k=1$.
This completes the proof of the argument.
\qed

As a side note, we remark that Volterra's function $V$ is more pathological
than the $\lambda$-Cantor function $\phi_\lambda$ for $\lambda \in (0,1)$,
in that $V'$ is discontinuous on a set of positive measure
even though $V$ is differentiable everywhere and $V'$ is bounded.
Contrast this with the fact that $\phi_\lambda'$  is also discontinuous on a set of positive measure,
but $\phi_\lambda$ is differentiable just almost everywhere, not everywhere.
In fact, there even exists a more pathological function $W : (0,1) \to \bR$ such that
$W$ is differentiable everywhere and $W'$ is bounded, but $W'$ is discontinuous almost everywhere
\cite[Exercise 5:5.5]{BrucknerBT97}.
Certainly, $W$ is an another example for \cref{prop:cont-ae-diff-not-pap}:
it is continuous and differentiable almost everywhere, but not PAP.

\subsection{Interior and subinterior of analytic partition}

\begin{definition}[Interior of Analytic Partition]
  \label{def:intrr-ap}
  Let $A=\{A^i\}_{i \in [I]}$ be an analytic partition of $\cX$.
  The {\bf interior} of $A$, $\intrr{A}$,
  is defined by the following open subset of $\bR^n$:
  \[\intrr{A} = \bigcup_{i \in [I]} \intrr{A^i; \bR^n}, \] %% = \bigcup_{i \in [I]} \intrr{A^i; \cX},\]
  where $\intrr{A^i; \bR^n}$ denotes the largest open subset of $\bR^n$ that is included in $A^i$.
  %% The second equality holds,
  %% since for any $S \subseteq \cX$, $S$ is open in $\cX$ iff $S$ is open in $\bR^n$.
\end{definition}

\begin{definition}[Subinterior of Analytic Partition]
  \label{def:subintrr}
  Let $A=\{A^i\}_{i \in [I]}$ be an analytic partition of $\cX$ with $I \in (\bZ_{>0} \cup \{\infty\})$.
  Suppose that a subset $\cX' \subseteq \cX$
  and a partition $B=\{B^{t}\}_{t \in [T]}$ of $\cX'$ satisfy the following conditions:
  \begin{compactenum}[(i)]
  \item $T \in (\bZ_{>0} \cup \{\infty\})$.
  \item For all $t \in [T]$, $B^{t}$ is {\bf subanalytic}.
    That is, for all $t \in [T]$,
    there exist $J_{t}, L_{t} \in \bZ_{\geq 0}$ and analytic functions
    $g^+_{t,j} : \cX^+_{t,j} \to \bR$ and
    $g^-_{t,l} : \cX^-_{t,l} \to \bR$
    over open domains $\cX^+_{t,j},\cX^-_{t,l} \subseteq \bR^n$
    ($j \in [J_t]$, $l \in [L_t]$) such that:
          $\cX^-_{t,l}$ is \emph{connected} in $\bR^n$
    and $(g^-_{t,l})^{-1}(\{0\}) \neq \cX^-_{t,l}$\; for all $l \in [L_t]$;
    and
    %% \begin{align}
    %%   \label{eq:subintrr}
    %%   \hspace{-2em}
    %%   B^{t} = \{ x \in \bR^n \mid \forall \langle j,l \rangle \in [J_{t}] \times [L_{t}].\,
    %%   (x \in \cX^+_{t,j} \land
    %%   g^+_{t,j}(x) > 0) \land
    %%   (x \in \cX^-_{t,l} \land
    %%   g^-_{t,l}(x) < 0) \}.
    %% \end{align}
    \begin{align}
      \label{eq:subintrr}
      B^t &=
      \Bigl( \bigcap_{j \in [J_t]}  (g^+_{t,j})^{-1}(\bR_{>0}) \Bigr) %% \cX^+_{t,j} \cap
      \cap
      \Bigl( \bigcap_{l \in [L_t]}  (g^-_{t,l})^{-1}(\bR_{<0}) \Bigr). %% \cX^-_{t,l} \cap
    \end{align}
  \item
    Let  $C^t \subseteq \bR^n$ be the set defined as follows:
    %% \[
    %% \hspace{-3.2em}
    %% C^{t} = \{ x \in \bR^n \mid \forall \langle j,l \rangle \in [J_{t}] \times [L_{t}].\,
    %% (x \in \cX^+_{t,j} \land
    %% g^+_{t,j}(x) > 0) \land
    %% (x \in \cX^-_{t,l} \land
    %% g^-_{t,l}(x) \leq 0) \}.
    %% \]
    \begin{align*}
      C^t &=
      \Bigl( \bigcap_{j \in [J_t]} (g^+_{t,j})^{-1}(\bR_{>0}) \Bigr) %% \cX^+_{t,j} \cap
      \cap
      \Bigl( \bigcap_{l \in [L_t]} (g^-_{t,l})^{-1}(\bR_{\leq 0}) \Bigr). %% \cX^-_{t,l} \cap
    \end{align*}
    Then, $C=\{C^t\}_{t \in [T]}$ is a finer partition of $\cX$ than $\{A^i\}_{i \in [I]}$.
    That is, $C$ is a partition of $\cX$,
    and for all $t \in [T]$,
    $C^t \subseteq A^i$ for some $i \in [I]$.
    %% for all $i \in [I]$,
    %% $A^i$ is a union of some elements in $\{C^{t}\}_{t \in [T]}$;
    %% %% $\cX = \bigcup_{t \in [T]} C^t$,
  \end{compactenum}
  We call the set $\cX'$ a {\bf subinterior} of $A$,
  and the partition $B$ a {\bf subanalytic} partition of $\cX'$.
  We use $\subintrr{A}$ to denote the set of all subinteriors of $A$.
\end{definition}

\begin{lemma}%[Properties of (Sub)Interior of Analytic Partition]
  \label{lem:subintrr-ap}
  Let $A = \{A^i\}_{i \in [I]}$ be an analytic partition of $\cX$
  with $I \in (\bZ_{>0} \cup \{\infty\})$.
  Then, $\subintrr{A} \neq \varnothing$.
  Furthermore, for any $\cX' \in \subintrr{A}$, the following hold:
  \begin{compactenum}[(a)]
  \item
    There is an analytic partition $B=\{B^t\}_{t \in [T]}$ of $\cX'$
    with $T \in (\bZ_{>0} \cup \{\infty\})$ and $\intrr{B} = \cX'$.
  \item 
    $\cX' \subseteq \intrr{A}$ and $\cX'$ is open in $\bR^n$.
  \item 
    $\cX \setminus \cX'$ is contained in some measure-zero set.
  \item 
    $\cX \setminus \intrr{A}$ is contained in some measure-zero set.
  \end{compactenum}
\end{lemma}
\begin{myproof}
  We first prove that $\subintrr{A} \neq \varnothing$.
  Consider any $i \in [I]$.
  Since $A^i$ is analytic,
  there exist $J_i, L_i \in \bZ_{>0}$
  and analytic functions
  $g^+_{i,j} : \cX^+_{i,j} \to \bR$ and
  $g^-_{i,l} : \cX^-_{i,l} \to \bR$
  over open domains $\cX^+_{i,j},\cX^-_{i,l} \subseteq \bR^n$
  ($j \in [J_i]$, $l \in [L_i]$) such that
  \[
  A^i =
  \Bigl( \bigcap_{j \in [J_i]} (g^+_{i,j})^{-1}(\bR_{>0}) \Bigr) %% \cX^+_{i,j} \cap
  \cap
  \Bigl( \bigcap_{l \in [L_i]} (g^-_{i,l})^{-1}(\bR_{\leq 0}) \Bigr). %% \cX^-_{i,l} \cap
  \]
  We use the following fact:
  %% reference: https://math.stackexchange.com/a/275769
        any open set in $\bR^n$ is a union of countably many \emph{open balls} in $\bR^n$,
        thereby a union of countably many \emph{disjoint connected} open sets in $\bR^n$.
  For every $l \in [L_i]$,
  since $\cX^-_{i,l}$ is open in $\bR^n$,
  there exists a partition $\{\cX^-_{i,\langle l,t_l \rangle}\}_{t_l \in [\infty]}$ of $\cX^-_{i,l}$
  such that $\cX^-_{i,\langle l,t_l \rangle}$ is connected and open in $\bR^n$.
  %% \begin{align}
  %%   \label{eq:subintrr-ap-connected}
  %%   \cX^-_{i,l} = \bigcup_{t_l \in [\infty]} \cX^-_{i,\langle l,t_l \rangle}.
  %% \end{align}
  For each $\langle t_1, \ldots, t_{L_i} \rangle \in [\infty]^{L_i}$, let
  %% \begin{align*}
  %%   B^{i,\langle t_1, \ldots, t_{L_i} \rangle}
  %%   &= \{ x \in \bR^n \mid \forall \langle j,l \rangle \in [J_{i}] \times [L_{i}].\,
  %%   (x \in \cX^+_{i,j} \land
  %%   g^+_{i,j}(x) > 0) \land
  %%   (x \in \cX^-_{i,\langle l,t_l \rangle} \land
  %%   g^-_{i,\langle l,t_l \rangle}(x) < 0) \},
  %%   \\
  %%   C^{i,\langle t_1, \ldots, t_{L_i} \rangle}
  %%   &= \{ x \in \bR^n \mid \forall \langle j,l \rangle \in [J_{i}] \times [L_{i}].\,
  %%   (x \in \cX^+_{i,j} \land
  %%   g^+_{i,j}(x) > 0) \land
  %%   (x \in \cX^-_{i,\langle l,t_l \rangle} \land
  %%   g^-_{i,\langle l,t_l \rangle}(x) \leq 0) \},
  %% \end{align*}
  \begin{align*}
    B^{i,\langle t_1, \ldots, t_{L_i} \rangle}
    &=
    \Bigl( \bigcap_{j \in [J_{i}]}  (g^+_{i,j})^{-1}(\bR_{>0}) \Bigr) %% \cX^+_{i,j} \cap
    \cap
    \Bigl( \bigcap_{l \in [L_{i}]} %% \cX^-_{i,\langle l,t_l \rangle} \cap
    (g^-_{i,\langle l,t_l \rangle})^{-1}(\bR_{<0}) \Bigr)
    \\
    C^{i,\langle t_1, \ldots, t_{L_i} \rangle}
    &=
    \Bigl( \bigcap_{j \in [J_{i}]}  (g^+_{i,j})^{-1}(\bR_{>0}) \Bigr) %% \cX^+_{i,j} \cap
    \cap
    \Bigl( \bigcap_{l \in [L_{i}]} %% \cX^-_{i,\langle l,t_l \rangle} \cap
    (g^-_{i,\langle l,t_l \rangle})^{-1}(\bR_{\leq 0}) \Bigr)
  \end{align*}
  where
  $g^-_{i,\langle l,t_l \rangle} : \cX^-_{i,\langle l,t_l \rangle} \to \bR$ denotes
  the restriction of $g^-_{i,l}$ to $\cX^-_{i,\langle l,t_l \rangle}$.
  Here, if $(g^-_{i,\langle l,t_l \rangle})^{-1}(\{0\}) = \cX^-_{i,\langle l,t_l \rangle}$ for some $i, l, t_l$,
  then we set $g^-_{i,\langle l,t_l \rangle}$ to the constant function $-1$
  on the domain $\cX^-_{i,\langle l,t_l \rangle}$.
  Then, every $g^-_{i,\langle l,t_l \rangle}$ is analytic
  on its connected open domain $\cX^-_{i,\langle l,t_l \rangle}$,
  and $(g^-_{i,\langle l,t_l \rangle})^{-1}(\{0\}) \neq \cX^-_{i,\langle l,t_l \rangle}$.
  Finally, let
  \begin{align*}
    \cX' &= \bigcup_{i \in [I], \;\langle t_1,\ldots,t_{L_i} \rangle \in [\infty]^{L_i}}
    B^{i,\langle t_1,\ldots,t_{L_i} \rangle}
    \\
    B &= \{B^{i,\langle t_1,\ldots,t_{L_i} \rangle} \mid
    i \in [I],\; \langle t_1,\ldots,t_{L_i} \rangle \in [\infty]^{L_i}\}
    \\
    C &= \{C^{i,\langle t_1,\ldots,t_{L_i} \rangle} \mid
    i \in [I],\; \langle t_1,\ldots,t_{L_i} \rangle \in [\infty]^{L_i}\}.
  \end{align*}
  Then, $\cX'$ is a subinterior of $A$, and
  $B$ is a subanalytic partition of $\cX'$,
  because of the following:
  \begin{compactitem}
  \item
    $B$ is a partition of $\cX'$.
  \item
    $\{ \langle i, t_1,\ldots,t_{L_i} \rangle \mid
    i \in [I],\; \langle t_1,\ldots,t_{L_i} \rangle \in [\infty]^{L_i}\}$
    is a countable set.
  \item
    For all $i \in [I]$ and $\langle t_1,\ldots,t_{L_i} \rangle \in [\infty]^{L_i}$,
    $B^{i,\langle t_1,\ldots,t_{L_i} \rangle}$ is subanalytic.
  \item
    $C$ is a finer partition of $\cX$ than $A$.
    This holds because
    $\{A^i\}_{i \in [I]}$ is a partition of $\cX$,
    and $\{C^{i, \langle t_1, \ldots, t_{L_i} \rangle}\}
    _{\langle t_1, \ldots, t_{L_i} \rangle \in [\infty]^{L_i}}$
    is a partition of $A^i$ for all $i \in [I]$,
    by its construction.
  \end{compactitem}
  This completes the proof that $\subintrr{A} \neq \varnothing$.

  We now prove the remaining claims.
  Let $\cX' \in \subintrr{A}$
  and $B=\{B^t\}_{t \in [T]}$ be a subanalytic partition of $\cX'$
  that satisfies the equations in \cref{def:subintrr}.
  %% We remark that the some parts of the below proof
  %% extend the proof of Theorem~1 in~\cite{ZhouGKRYW19}.

  Proof of (a).
  By the definition of subanalytic partition,
  and since $h(x) < 0 \Longleftrightarrow -h(x) > 0$ for any function $h$ and input $x$,
  $B$ is an analytic partition of $\cX'$ with $T \in (\bZ_{>0} \cup \{\infty\})$.
  We argue that for any $t \in [T]$, $B^t$ is open in $\bR^n$.
  Recall the equation~\eqref{eq:subintrr}:
  \begin{align*}
    B^t &=
    \Bigl( \bigcap_{j \in [J_t]}  (g^+_{t,j})^{-1}(\bR_{>0}) \Bigr) %% \cX^+_{t,j} \cap
    \cap
    \Bigl( \bigcap_{l \in [L_t]}  (g^-_{t,l})^{-1}(\bR_{<0}) \Bigr). %% \cX^-_{t,l} \cap
  \end{align*}
  Since $g^+_{t,j} : \cX^+_{t,j} \to \bR$ and $g^-_{t,l} : \cX^-_{t,l} \to \bR$ are continuous,
  and $\bR_{>0}$ and $\bR_{<0}$ are open in $\bR$,
  we have that $(g^+_{t,j})^{-1}(\bR_{>0})$ and $(g^-_{t,l})^{-1}(\bR_{<0})$ are open
  in $\cX^+_{t,j}$ and $\cX^-_{t,l}$, respectively, by the definition of continuity.
  Furthermore, since $\cX^+_{t,j}$ and $\cX^-_{t,l}$ are open in $\bR^n$,
  we have that $(g^+_{t,j})^{-1}(\bR_{>0})$ and $(g^-_{t,l})^{-1}(\bR_{<0})$ are open in $\bR^n$ as well.
  Since any finite intersection of open subsets is again open,
  $B^t$ is open in $\bR^n$.
  Hence, \[\intrr{B} = \bigcup_{t \in [T]} \intrr{B^t;\bR^n} = \bigcup_{t \in [T]} {B^t} = \cX'.\]
  
  Proof of (b).
  We continue the proof from (a).
  Since $B^t$ is open in $\bR^n$, and $B^t \subseteq A^i$ for some $i \in [I]$
  (by the definition of subanalytic partition),
  we have $B^t \subseteq \intrr{A^i; \bR^n}$.
  From this, we obtain 
  \[\cX' = \bigcup_{t \in [T]} B^t \subseteq \bigcup_{i \in [I]} \intrr{A^i;\bR^n} = \intrr{A}.\]
  Moreover, since any union of open sets is again open,
  $\cX' = \bigcup_{t \in [T]} B^t$ is open in $\bR^n$.

  Proof of (c).
  For this, we use the following theorem~\cite{Mityagin15}:
  for any open connected $U \subseteq \bR^n$
  and analytic function $f: U \to \bR$,
  the zero set $\{x \in U \mid f(x)=0\}$ of $f$ is either $U$
  or contained in some measure-zero set.
  Observe that
  \begin{align*}
    \cX \setminus \cX'
    &= \Bigl(\bigcup_{t \in [T]} C^t \Bigr) \setminus \Bigl( \bigcup_{t \in [T]} B^t\Bigr)
    \\
    &= \bigcup_{t \in [T]} \Bigl( C^t \setminus B^t \Bigr)
    \\
    & \subseteq
    \bigcup_{t \in [T]}
    \Bigl( \bigcap_{l \in [L_t]}  (g^-_{t,l})^{-1}(\bR_{\leq 0})
    \setminus
    \bigcap_{l \in [L_t]}  (g^-_{t,l})^{-1}(\bR_{< 0}) \Bigr)
    \\
    & \subseteq
    \bigcup_{t \in [T]} \bigcup_{l \in [L_t]}
    \Bigl( (g^-_{t,l})^{-1}(\bR_{\leq 0}) \setminus (g^-_{t,l})^{-1}(\bR_{< 0}) \Bigr)
    \\
    & = 
    \bigcup_{t \in [T]} \bigcup_{l \in [L_t]} (g^-_{t,l})^{-1}(\{0\}).
  \end{align*}
  Since each $g^-_{t,l}$ is analytic, and not everywhere-zero,
  on its connected open domain $\cX^-_{t,l}$
  (by the definition of subanalytic partition),
  the above theorem and equation imply that
  $(g^-_{t,l})^{-1}(\{0\})$ is contained in some measure-zero set.
  Since any countable union of measure-zero sets has measure zero,
  $\cX \setminus \cX'$ is contained in some measure-zero set.
     
  Proof of (d).
  This follows immediately from (b) and (c). %% (in particular, $\cX' \subseteq \intrr{A}$)
  \qed
\end{myproof}

\subsection{Proofs of \cref{prop:pap-ae-diff} %% Proposition
  and \cref{prop:intr-deriv-coincidence} (part I)} %% Proposition
\label{sec:proof-prop-pap-ae-diff}

We remind the reader that the notation $D(f)(x)$ means the standard derivative of $f$ at $x$.

\begin{definition}[Interior of PAP Representation]
  \label{def:intrr-pap-repr}
  Let $\gamma = \{\langle A^i,f^i \rangle\}_{i \in [I]}$ be a PAP representation from $\cX$ to $\cY$.
  The {\bf interior} and {\bf subinterior} of $\gamma$ are defined by:
  \begin{align*}
    \intrr{\gamma} &= \intrr{\{A^i\}_{i \in [I]}}, &
    \subintrr{\gamma} &= \subintrr{\{A^i\}_{i \in [I]}}.
  \end{align*}
\end{definition}

\begin{lemma}
  \label{lem:deriv-pap-func}
  Let $f : \cX \to \cY$ be a PAP function,
  $\gamma$ be a PAP representation of $f$,
  and $k \in \bZ_{\geq 0}$.
  Then, for all $x \in \intrr{\gamma}$,
  $f$ has the $k$-th order standard derivative at $x$.
  Furthermore, 
  the derivative agrees with the $k$-th order intensional derivative of $\gamma$ at $x$:
  \[ D^{(k)}(f)(x) = \eval{D^{(k)}(\gamma)}(x)
  \quad\text{for all $x \in \intrr{\gamma}$},\]
  where $F^{(k)}$ denotes the $k$-time composition of the operator $F$.
\end{lemma}
\begin{myproof}
  Consider any $x \in \intrr{\gamma}$.
  By the definition of $\intrr{\gamma}$, we have $x \in \intrr{A^i; \bR^n}$ for some $i \in [I]$.
  So there exists an open neighbourhood $U \subseteq \bR^n$ of $x$ such that $U \subseteq A^i$.
  Since $\gamma$ is a representation of $f$ and $U \subseteq A^i$, we have $f = f^i$ on $U$ for the $i$-th component
  function of $\gamma$.
  Now focus on $f^i$.
  Since $f^i$ is analytic on $U$ (due to $\gamma$ being PAP),
  $f^i$ is infinitely differentiable on $U$
  and, in particular, has the $k$-th order standard derivative at $x$, namely $D^{(k)}(f^i)(x)$.
  By the definition of intensional derivative, %% $D^{(k)}(\gamma)$ and $\eval{\cdot}$,
  $D^{(k)}(f^i)(x) = \eval{D^{(k)}(\gamma)}(x)$.
  Since $x \in U$, $U$ is open in $\bR^n$, and $f=f^i$ on $U$,
  we obtain that $D^{(k)}(f)(x) = D^{(k)}(f^i)(x) = \eval{D^{(k)}(\gamma)}(x).$
  \qed
\end{myproof}

\begin{proposition}
  \label{prop:deriv-pap-func}
  Let $f : \cX \to \cY$ be a PAP function and $k \in \bZ_{\geq 0}$.
  Then, $f$ has the $k$-th order standard derivative almost everywhere.
  Moreover, the first-order standard derivative $Df$ agrees with
  any first-order intensional derivative $\mathit{df} \in \ideriv{f}$
  almost everywhere.
\end{proposition}
\begin{myproof}
  By \cref{lem:subintrr-ap}(d),
  the interior of any PAP representation of $f$ has the full measure in $\cX$.
  Hence, \cref{lem:deriv-pap-func} implies the first claim.
  For the second claim, let $\mathit{df} \in \ideriv{f}$.
  By the definition of $\ideriv{f}$,
  there exists a PAP representation $\gamma$ of $f$
  such that $\mathit{df} = \eval{D\gamma}$.
  Applying \cref{lem:deriv-pap-func} to $f$, $\gamma$, and $k=1$
  gives the second claim,
  since $\intrr{\gamma}$ has the full measure in $\cX$.
  \qed
\end{myproof}

\subsection{Proof of \cref{prop:intr-deriv-coincidence} (part II)} %% Proposition
%% \subsection{PAP Function and Its Standard/Intensional Derivative}
\label{sec:proof-prop-intr-deriv-coincidence}

\begin{lemma}
  \label{lem:pap-funcs-eq-deriv}
  Let $f : \cX_f \to \cY$ and $g : \cX_g \to \cY$ be PAP functions,
  and $\gamma_f$ and $\gamma_g$ be their PAP representations.
  If $f(x) = g(x)$ for all $x \in \intrr{\gamma_f} \cap \intrr{\gamma_g}$,
  then %% $(Df)(x)$ and $(Dg)(x)$ exist and
  \[D(f)(x) = D(g)(x) \quad\text{for all $x \in \intrr{\gamma_f} \cap \intrr{\gamma_g}$},\]
  where 
        both sides are well-defined for each $x$.
\end{lemma}
\begin{myproof}
  Let $U = \intrr{\gamma_f} \cap \intrr{\gamma_g}$, and consider any $x \in U$.
  Since $x \in U$ and $U$ is open in $\bR^n$,
  we have $D(f)(x) = D(g)(x)$ if both sides are well-defined.
  Indeed, they are well-defined by \cref{lem:deriv-pap-func} with $k=1$,
  since $\gamma_f$ and $\gamma_g$ are PAP representations of $f$ and $g$, respectively.
  \qed
\end{myproof}

\begin{lemma}
  \label{lem:pap-repr-intr-deriv}
  Let $f : \cX \to \cY$ be a PAP function.
  Then, for any intensional derivative $\mathit{df} \in \ideriv{f}$,
  there exists a PAP representation $\gamma_{\mathit{df}}$ of $\mathit{df}$ such that
  \[\mathit{df}(x) = D(f)(x) \quad\text{for all $x \in \intrr{\gamma_{\mathit{df}}}$.}\]
\end{lemma}
\begin{myproof}
  By the definition of $\ideriv{f}$,
  there exists a PAP representation $\gamma_f$ of $f$ such that $\mathit{df} = \eval{D\gamma_f}$.
  By \cref{lem:deriv-pap-func} with $k=1$,
  we have $D(f)(x) = \eval{D \gamma_f}(x) = \mathit{df}(x)$ for all $ x \in \intrr{\gamma_f}$.
  Let $\gamma_{\mathit{df}} = D\gamma_f$.
  Since $\mathit{df} = \eval{D\gamma_f} = \eval{\gamma_{\mathit{df}}}$ and $\gamma_f$ is PAP,
  $\gamma_{\mathit{df}}$ is a PAP representation of $\mathit{df}$.
  Moreover, by the definition of $D\gamma_f$,
  we have $\intrr{\gamma_{\mathit{df}}} = \intrr{\gamma_f}$.
  Hence, the claim holds with $\gamma_{\mathit{df}}$.
  \qed
\end{myproof}

\begin{definition}[Refinement of Representation]
  Let $\gamma_f = \{\langle A^i, f^i \rangle\}_{i \in [I]}$
  be a representation of a function from $\cX_f$ to $\cY$,
  and $B = \{B^j\}_{j \in [J]}$ be a partition of $\cX_g$.
  The {\bf refinement} of $\gamma_f$ with $B$ is defined by:
  \[\refine{\gamma_f; B} = \{\langle A^i \cap B^j, f^i \rangle\}_{\langle i,j \rangle \in [I] \times [J]}.\]
  Moreover, for any representation $\gamma_g = \{\langle C^l, g^l \rangle\}_{l \in [L]}$
  of a function from $\cX_g$ to $\cZ$,
  the {\bf refinement} of $\gamma_f$ with $\gamma_g$ is defined by:
  \[\refine{\gamma_f; \gamma_g} = \refine{\gamma_f; \{C^l\}_{l \in [L]}}.\]
\end{definition}

\begin{lemma}
  \label{lem:pap-repr-refine}
  Let $f : \cX_f \to \cY$ be a PAP function,
  $\gamma$ be a PAP representation of $f$,
  and $B = \{B^j\}_{j \in [J]}$ be an analytic partition of $\cX_g$
  with $J \in (\bZ_{>0} \cup \{\infty\})$.
  Let $\gamma' = \refine{\gamma; B}$.
  Then, $\gamma'$ is a PAP representation of $f |_{\cX_f \cap \cX_g}$
  with
  \[
  \intrr{\gamma'} = \intrr{\gamma} \cap \intrr{B}.
  \]
\end{lemma}
\begin{myproof}
  Let $\gamma = \{\langle A^i, f^i \rangle\}_{i \in [I]}$.
  Since $\gamma$ is PAP and $B$ is an analytic partition,
  $\{A^i \cap B^j\}_{\langle i,j \rangle \in [I] \times [J]}$ is an analytic partition.
  Also, since $[J]$ is countable, $[I] \times [J]$ is also countable.
  Thus, $\gamma'$ is PAP.
  Since \[\bigcup_{\langle i,j \rangle \in [I] \times [J]} (A^i \cap B^j)
  = (\bigcup_{i \in [I]} A^i) \cap (\bigcup_{j \in [J]} B^j)
  = \cX_f \cap \cX_g,\]
  $\gamma'$ is a representation of ${f|_{\cX_f \cap \cX_g}}$.
  Finally, we obtain the last claim as follows:
  \begin{align*}
    \intrr{\gamma'}
    &= \bigcup_{\langle i,j \rangle \in [I] \times [J]} \intrr{A^i \cap B^j; \bR^n}
    \\
    &= \bigcup_{\langle i,j \rangle \in [I] \times [J]} \intrr{A^i; \bR^n} \cap \intrr{B^j; \bR^n}
    \\
    &= \Bigl(\bigcup_{i \in [I]} \intrr{A^i; \bR^n}\Bigr) \cap
    \Bigl(\bigcup_{j \in [J]} \intrr{B^j; \bR^n}\Bigr)
    \\
    &= \intrr{\gamma} \cap \intrr{B}.
  \end{align*}
  For the second equality, we use the following fact:
  $\intrr{S_1 \cap S_2; X} = \intrr{S_1; X} \cap \intrr{S_2; X}$
  for any $S_1, S_2 \subseteq X$.
  \qed
\end{myproof}

%% \begin{definition}
%%   Let $f : \cX \to \cY$ and $k \geq 1$.
%%   Define a finite extension of $D^{(k)}f : \cX \to \bR^{m \times n^k} \cup \{\bot\}$,
%%   denoted by $\overline{D^{(k)}f} : \cX \to \bR^{m \times n^k}$, as follows:
%%   \wl{TODO.}
%% \end{definition}

%% \begin{lemma}
%%   There exists a collection of intensional representations
%%   $\{ \overline{\gamma^k} \in \irepr{\overline{D^{(k)}f}} \,|\, k \geq 1 \}$
%%   such that ${\rm int}(\overline{\gamma^k}) = {\rm int}(\overline{\gamma^{k'}})$
%%   for any $k, k' \geq 1$.
%% \end{lemma}

%% \begin{proposition}
%%   Let $k \geq 0$ and $f : \cX \to \cY$.
%%   Then, for any $\mathit{df}^k \in \iderivk{k}f$,
%%   there exist $d\gamma^k \in \irepr{\mathit{df}^k}$ and $\overline{\gamma^k} \in \irepr{\overline{D^{(k)}f}}$
%%   such that $\mathit{df}^k(x) = \overline{D^{(k)}f}(x)$
%%   for all $x \in {\rm int}(d\gamma^k) \cap {\rm int}(\overline{\gamma^k})$.
%% \end{proposition}

\begin{lemma}
  \label{lem:intr-deriv-coincidence-pt1}
  Let $f : \cX \to \cY$ be a PAP function.
  Consider any PAP representation $\gamma$ of $f$,
  and any subinterior $\cX' \in \subintrr{\gamma}$.
  Then, $\cX' \subseteq \cX$ is open in $\bR^n$ and
  $\cX \setminus \cX'$ is contained in a measure-zero set.
  Moreover, for all $k \in \bZ_{\geq 0}$,
  $D^{(k)}(f|_{\cX'})$ is a total function on $\cX'$,
  and there exists a PAP representation $\gamma_D^k$ of $D^{(k)}(f|_{\cX'})$
  such that $\intrr{\gamma_D^k} = \cX'$.
\end{lemma}
\begin{myproof}
  Let $k \in \bZ_{\geq 0}$.
  By \cref{lem:subintrr-ap}(b) and~\ref{lem:subintrr-ap}(c),
  $\cX' \subseteq \intrr{\gamma} \subseteq \cX$ is open in $\bR^n$ and
  $\cX \setminus \cX'$ is contained in a measure-zero set.
  This proves the first claim.
  Since $\cX' \subseteq \intrr{\gamma}$,
  \cref{lem:deriv-pap-func} implies that
  $D^{(k)}(f)(x)$ exists for all $x \in \cX'$.
  Since $\cX'$ is open in $\bR^n$,
  $D^{(k)}(f|_{\cX'})(x) = D^{(k)}(f)(x)$ for all $x \in \cX'$.
  This proves the second claim that
  $D^{(k)}(f|_{\cX'})$ is a total function on $\cX'$.

  We now prove the last claim.
  By \cref{lem:subintrr-ap}(a),
  there exists an analytic partition $B=\{B^j\}_{j \in [J]}$ of $\cX'$ such that
  $J \in (\bZ_{>0} \cup \{\infty\})$ and $\intrr{B} = \cX'$.
  Let $\gamma' = \refine{\gamma; B}$.
  By \cref{lem:pap-repr-refine}, $\gamma'$ is a PAP representation of $f|_{\cX'}$
  with $\intrr{\gamma'} = \intrr{\gamma} \cap \intrr{B} = \cX'$.
  Consider \[\gamma_D^k = D^{(k)}(\gamma').\]
  We show that it satisfies the last claim.
  Since $f|_{\cX'} : \cX' \to \cY$ is a PAP function with a PAP representation $\gamma'$,
  \cref{lem:deriv-pap-func} implies that
  $D^{(k)}(f|_{\cX'})(x) = \eval{D^{(k)}(\gamma')}(x) = \eval{\gamma_D^k}(x)$
  for all $x \in \intrr{\gamma'} = \cX'$.
  Hence, $\gamma_D^k$ is a representation of $D^{(k)}(f|_{\cX'})$.
  The rest of the claim also holds as follows:
  $\gamma_D^k$ is PAP since $\gamma'$ is PAP;
  and $\intrr{\gamma_D^k} = \intrr{\gamma'} = \cX'$
  by the definition of $D(\gamma')$.
  \qed
\end{myproof}

\begin{lemma}
  \label{lem:intr-deriv-coincidence-pt2}
  Let $f : \cX \to \cY$ be a PAP function 
  and $\cX'\subseteq \cX$ be a set described in \cref{lem:intr-deriv-coincidence-pt1},
  which has full measure in $\cX$.
  Consider any $k \in \bZ_{\geq 0}$.
  Then, for any $\mathit{df}^k \in \iderivk{k}f$,  % $k$-th order intensional derivative
  there exists a PAP representation $\gamma_d^k$ of $\mathit{df}^k$ 
  such that
  \[\mathit{df}^k(x) = D^{(k)}(f)(x)
  \quad
  \text{for all $x \in \intrr{\gamma_d^k} \cap \cX'$.}\]
\end{lemma}
\begin{myproof}
  The proof proceeds by induction on $k$.
  For $k=0$, we have $\mathit{df}^k = D^{(k)}(f) = f$.
  So any PAP representation $\gamma_d^k$ of $\mathit{df}^k=f$ satisfies the claim.
  Now suppose $k>0$.
  By the definition of $\iderivk{k}{f}$,
  there exists $\mathit{df}^{k-1} \in \iderivk{k-1}{f}$ such that $\mathit{df}^k \in \ideriv{(\mathit{df}^{k-1})}$.
  We construct the desired PAP representation $\gamma_d^k$ as follows.
  First, focus on $\mathit{df}^{k-1} \in \iderivk{k-1}{f}$.
  By the induction hypothesis on $k-1$ for $\mathit{df}^{k-1}$,
  there exists a PAP representation $\gamma_d^{k-1}$ of $\mathit{df}^{k-1}$ such that
  \begin{align*}
    \mathit{df}^{k-1}(x) &= D^{(k-1)}(f)(x)
    && \text{for all $x \in \intrr{\gamma_d^{k-1}} \cap \cX'$.}
    \intertext{
      By \cref{lem:intr-deriv-coincidence-pt1},
      $D^{(k-1)}(f|_{\cX'})$ is a total function on $\cX'$
      and there exists a PAP representation $\gamma_D^{k-1}$ of $D^{(k-1)}(f|_{\cX'})$
      such that $\intrr{\gamma_D^{k-1}} = \cX'$.
      Since $\cX'$ is open in $\bR^n$ by \cref{lem:intr-deriv-coincidence-pt1},
      $D^{(k-1)}(f) = D^{(k-1)}(f|_{\cX'})$ over $\cX'$.
      Combining the two results gives that
    }
    \mathit{df}^{k-1}(x) &= D^{(k-1)}(f|_{\cX'})(x)
    && \text{for all $x \in \intrr{\gamma_d^{k-1}} \cap \intrr{\gamma_D^{k-1}}$,}
    \intertext{
      where
      $\gamma_d^{k-1}$ and $\gamma_D^{k-1}$ are
      PAP representations of ${\mathit{df}^{k-1}}$ and ${D^{(k-1)}(f|_{\cX'})}$, respectively.
      By \cref{lem:pap-funcs-eq-deriv} applied to this result,
      we obtain
    }
    D(\mathit{df}^{k-1})(x) &= D^{(k)}(f|_{\cX'})(x)
    && \text{for all $x \in \intrr{\gamma_d^{k-1}} \cap \intrr{\gamma_D^{k-1}}$.}
    \intertext{
      Since $D^{(k)}(f|_{\cX'}) = D^{(k)}(f)$ over $\cX'$
      (as $\cX'$ is open in $\bR^n$),
      and $\intrr{\gamma_D^{k-1}} = \cX'$,
      we have
    }
    D(\mathit{df}^{k-1})(x) &= D^{(k)}(f)(x)
    && \text{for all $x \in \intrr{\gamma_d^{k-1}} \cap \cX'$.}
    \intertext{
      Next, focus on $\mathit{df}^k \in \ideriv{(\mathit{df}^{k-1})}$.
      By \cref{lem:pap-repr-intr-deriv} applied to $\mathit{df}^k$,
      there exists a PAP representation $\gamma_d^{\prime k}$ of $\mathit{df}^k$
      such that
    }
    \mathit{df}^k(x) &= D(\mathit{df}^{k-1})(x)
    && \text{for all $x \in \intrr{\gamma_d^{\prime k}}$.}
    \intertext{
      Combining the last two equations, we obtain
    }
    \mathit{df}^{k}(x) &= D^{(k)}(f)(x)
    && \text{for all $x \in \intrr{\gamma_d^{\prime k}} \cap \intrr{\gamma_d^{k-1}} \cap \cX'$.}
    \intertext{
      Now let $\gamma_d^k = \refine{\gamma_d^{\prime k}; \gamma_d^{k-1}}$.
      By \cref{lem:pap-repr-refine} applied to $\gamma_d^k$,
      we have that $\gamma_d^k$ is a PAP representation of $\mathit{df}^k$ with
      $\intrr{\gamma_d^k} = \intrr{\gamma_d^{\prime k}} \cap \intrr{\gamma_d^{k-1}}$.
      From this, we obtain the desired claim:
    }
    \mathit{df}^{k}(x) &= D^{(k)}(f)(x)
    && \text{for all $x \in \intrr{\gamma_d^{k}} \cap \cX'$.}
  \end{align*}
  \qed
\end{myproof}

\begin{proposition}
  \label{prop:intr-deriv-coincidence-full}
  Let $f : \cX \to \cY$ be a PAP function and $k \in \bZ_{\geq 0}$.
  Then, any $k$-th order intensional derivative $\mathit{df}^k \in \iderivk{k}f$
  satisfies the following:
  \[\mathit{df}^k(x) = D^{(k)}(f)(x) \quad\text{for almost all $x \in \cX$.}\]
\end{proposition}
\begin{myproof}
  The claim follows from \cref{lem:intr-deriv-coincidence-pt2}
  and the following:
  $\cX'$ and $\intrr{\gamma_d^k}$ described in \cref{lem:intr-deriv-coincidence-pt2}
  have the full measure in $\cX$,
  by \cref{lem:intr-deriv-coincidence-pt2} and \cref{lem:subintrr-ap}(d).
  \qed
\end{myproof}

%auto-ignore

\subsection{Additional property on PAP functions} %% and intensional derivatives}
\label{sec:proof-rest}

\begin{proposition}
  \label{prop:pap-ndiff-countable}
  Let $f : \cX \to \cY$ be a PAP function,
  and $k \in \bZ_{>0}$.
  Let $N_{f,k} \subseteq \cX$ be the set defined by
  $\{ x \in \cX \mid D^{(k)}(f)(x) \text{ is undefined} \}$.
  If $n=1$, then $N_{f,k}$ is countable.
  But if $n>1$, then it could be uncountable.
\end{proposition}
\begin{myproof}
  Suppose $n=1$.
  Recall the following two well-known results:
  (i) if $g : U \to \bR$ is an analytic function on an open interval $U \subseteq \bR$
  and is not everywhere-zero,
  then its zero set $Z_g = \{ x \in U \mid g(x) = 0\}$
  contains none of the limit points of $Z_g$
  \cite[Corollary 1.2.7]{KrantzP02};
  (ii) if $X$ is a second-countable space
  and $S \subseteq X$ contains none of the limit points of $S$,
  then $S$ is countable.
  %% ref: https://math.stackexchange.com/a/2323591
  Since $\bR$ is second-countable,
  every $g$ satisfying the assumption of (i) has at most countably many zeros.
  
  Now return to our claim.
  By the statement and proof of \cref{lem:subintrr-ap}(c) and \cref{lem:deriv-pap-func},
  there are countably many analytic functions $\{g_j : U_j \to \bR \}_j$
  defined over connected open subsets of $\bR$,
  such that every $g_j$ is not everywhere-zero and $N_{f,k} \subseteq \bigcup_j Z_{g_j}$.
  Since any connected open subset of $\bR$ is an open interval,
  $g_j$ satisfies the assumption of (i),
  and thus each $Z_{g_j}$ is countable by the above result.
  Hence, $N_{f,k}$ is countable, since a countable union of countable sets is countable.
  
  Suppose $n>1$.
  Consider $f: \bR^2 \to \bR$ defined by $f(x) = |x_1-x_2|$, and $k=1$.
  Then, $f$ is PAP, since the following is a PAP representation of $f$:
  \begin{align*}
    \{ \quad
    &\langle
    \{ x \in \bR^2 \mid x_1 > x_2 \}, x \in \bR^2 \longmapsto x_1-x_2
    \rangle,
    \\
    &\langle
    \{ x \in \bR^2 \mid x_1 = x_2 \}, x \in \bR^2 \longmapsto 0
    \rangle,
    \\
    &\langle
    \{ x \in \bR^2 \mid x_1 < x_2 \}, x \in \bR^2 \longmapsto x_2-x_1
    \rangle
    \quad \}.
  \end{align*}
  However, $N_{f,k} = \{\langle x,y \rangle \in \bR^2 \mid x=y\}$ is uncountable.
  \qed
\end{myproof}

\commentout{
  \hsy{Later if we write a journal version of this paper, we may include these results.}
  \wl{See {\tt note-fst-deriv.tex} for the omitted results.
    Theses include the following:
    proposition on PAP and local Lipschitzness,
    remark on summarising examples for each class of functions,
    proposition characterising intensional derivatives,
    remark on valid version of Claim 3 using PAP and intensional derivatives,
    and remark on two open questions.}
}

%auto-ignore

\showConfn{
%% ========== CONFN ==========
\section{Proofs for Results in \S\ref{sec:autodiff}}
\label{sec:proof-autodiff}
%% ========== CONFN ==========
}

\showConfn{
%% ========== CONFN ==========
%% \begin{proposition}
%% \label{prop:semantics-pap}
{\bf \cref{prop:semantics-pap}.\,}
{\it For every program $e$, its denotation $\sem{e}$ is a PAP function from $\bR^N$ to $\bR$.}
%% \end{proposition}
\begin{myproof}
        The proof is by induction on the structure of $e$. The cases of $e \equiv c$ and $e \equiv x_i$ follow from the fact that both constant functions and projections are PAP. For the case of $e \equiv \bar{\code{f}}(e_1,\ldots,e_n)$, we note two facts about PAP functions: the composition of two PAP functions is PAP (\cref{prop:fn-comp-PAP}); and for PAP functions $g_1,\ldots,g_n : \bR^N \to \bR$, the function $v \longmapsto \langle g_1(v),\ldots,g_n(v) \rangle$ of type $\bR^N \to \bR^n$ is PAP again,
mainly because any finite intersection of open sets is again open.
By these two facts, the claimed property of the proposition holds in this case. 
The only remaining case is $e \equiv (\code{if}~(e_1>0)~e_2~e_3)$.
By induction hypothesis, all of $\sem{e_1}$, $\sem{e_2}$, and $\sem{e_3}$ are PAP.
Let $\gamma_1=\{\langle A^i,f^i \rangle\}_{i \in [I]}$, $\gamma_2=\{\langle B^j,g^j \rangle\}_{j \in [J]}$,
and $\gamma_3=\{\langle C^k,h^k \rangle\}_{k \in [K]}$ be their PAP representations,
and define their conditional composition $\ifelse{\gamma_1}{\gamma_2}{\gamma_3}$ as follows:
%% For each $i \in [I]$, define $A^i_1 = \{v \in A^i \mid f^i(v) > 0\}$ and $A^i_0 = \{v \in A^i \mid f^i(v) \leq 0\}$. Also, for all $\langle i,j,k,l\rangle \in ([I]\times [J] \times [K] \times \{0,1\})$, let
%% $D^{\langle i,j,k,l \rangle} = A^i_l \cap B^j \cap C^k$ and 
%% \[
%%         t^{\langle i,j,k,l \rangle} : \dom(f^i) \cap \dom(g^j) \cap \dom(h^k) \to \bR,
%%         \qquad
%%         t^{\langle i,j,k,l \rangle}(v) = 
%%         \begin{cases}
%%                 g^j(v) & \text{if $l=1$}
%%                 \\
%%                 h^k(v) & \text{if $l=0$}.
%%         \end{cases}
%% \]
\begin{align*}
  \ifelse{\gamma_1}{\gamma_2}{\gamma_3}
  &= \rlap{$\{\langle E^{\langle i,j,k,l \rangle},t^{\langle i,j,k,l \rangle} \rangle\}_{\langle i,j,k,l \rangle \in ([I]\times [J] \times [K] \times \{0,1\})},$}
  \\
  E^{\langle i,j,k,l \rangle} &= A^i_l \cap B^j \cap C^k,
  &
  t^{\langle i,j,k,l \rangle} &= \text{if } (l=1) \text{ then } g^j \text{ else } h^k
  \qquad\qquad
\end{align*}
where $A^i_1 = \{v \in A^i \mid f^i(v) > 0\}$ and $A^i_0 = \{v \in A^i \mid f^i(v) \leq 0\}$.
Then, $\{E^{\langle i,j,k,l \rangle}\}_{\langle i,j,k,l \rangle}$ is an analytic partition of $\bR^N$, every $t^{\langle i,j,k,l \rangle}$ is an analytic function, and its domain is an open set containing $E^{\langle i,j,k,l \rangle}$. Thus,
%% $\gamma = \{\langle E^{\langle i,j,k,l \rangle},t^{\langle i,j,k,l\rangle} \rangle\}_{\langle i,j,k,l \rangle}$
$\ifelse{\gamma_1}{\gamma_2}{\gamma_3}$
is a PAP representation. Furthermore, its evaluation
%% $\eval{\gamma}$
$\eval{\ifelse{\gamma_1}{\gamma_2}{\gamma_3}}$
is equal to $\sem{\code{if}~(e_1>0)~e_2~e_3}$. Hence, the proposition holds in this case.
  \qed
\end{myproof}
%% ========== CONFN ==========
}

\showConfn{
%% ========== CONFN ==========
%% \begin{theorem}
%%   \label{thm:autodiff-correct-fst}
{\bf \cref{thm:autodiff-correct-fst}.\,}
{\it
  %% For every program $e$, we have $\semdiff{e} \in \ideriv{\sem{e}}$.
  If $\mytilde{D}\code{f} \in \ideriv{{\code{f}}}$\,
  %% for every primitive-function symbol $\bar{\code{f}}$ in the language,
  for all primitive functions $\bar{\code{f}}$,
  then $\semdiff{e} \in \ideriv{\sem{e}}$\, for all programs $e$.}
%% \end{theorem}
\begin{myproof}
        The proof is by induction on the structure of $e$. When $e \equiv c$, the trivial partition $\{\bR^N\}$ and the constant function $v\longmapsto c$ form a PAP representation of $\sem{e}$. The intensional derivative of this representation is $\{\langle \bR^N, v \longmapsto \vec{0}_{1 \times N}\rangle \}$, and its evaluation is $\semdiff{c}$, as claimed by the theorem. The other base case is $e \equiv x_i$. We use the trivial partition again with the projection function $v \longmapsto v_i$, and form a PAP representation of $\sem{e}$. The intensional derivative of this representation is $\{\langle \bR^N, v \longmapsto [ \vec{0}_{(i-1) \times 1}; \vec{1}_{1 \times 1}; \vec{0}_{(N-i) \times 1}]^\top \rangle\}$, and its evaluation is $\semdiff{x_i}$. 

        The next case is $e \equiv \bar{\code{f}}(e_1,\ldots,e_n)$. By induction hypothesis, $\semdiff{e_i}$ is an intensional derivative of  $\sem{e_i}$ for every $i \in [n]$. Let $g : \bR^N \to \bR^n$ and $\mathit{dg} : \bR^N \to \bR^{n \times N}$ be functions defined by $g(v) = \langle \sem{e_1}v,\ldots,\sem{e_n}v\rangle$ and $\mathit{dg}(v) = [ \semdiff{e_1}v;\ldots;\semdiff{e_n}v ]$ for all $v$. Then, $\mathit{dg}$ is an intensional derivative of $g$. Also, $\sem{\bar{\code{f}}(e_1,\ldots,e_n)} = {\code{f}} \circ g$. Therefore, by the chain rule for intensional derivative (\cref{prop:intr-deriv-chain-rule}), the function $v \longmapsto  (\mytilde{D}\code{f})(g(v)) \cdot \mathit{dg}(v)$ is an intensional derivative of $\sem{\bar{\code{f}}(e_1,\ldots,e_n)} = {\code{f}} \circ g$.
        Here we use the assumption that  %%a part of the first requirement that
        $\mytilde{D}\code{f}$ is an intensional derivative of $\code{f}$.
        Note that the function is equal to $\semdiff{\bar{\code{f}}(e_1,\ldots,e_n)}$. So, the theorem holds.

The last case is $e \equiv (\code{if}~(e_1>0)~e_2~e_3)$. By induction hypothesis, $\semdiff{e_i}$ is an intensional derivative of $\sem{e_i}$ for all $i \in [3]$.
Let $\gamma_1$, $\gamma_2$, and $\gamma_3$ be the PAP representations of
$\sem{e_1}$, $\sem{e_2}$, and $\sem{e_3}$ such that
\begin{equation}
  \label{thm:autodiff-correct-fst:eqn1}
  \eval{D\gamma_2} = \semdiff{e_2}
  \quad\text{and}\quad
  \eval{D\gamma_3} = \semdiff{e_3}.
\end{equation}
Let $\ifelse{\gamma_1}{\gamma_2}{\gamma_3}$ be
the conditionally composed representation defined in the proof of \cref{prop:semantics-pap}.
Then, it is a PAP representation of $\sem{\code{if}~(e_1>0)~e_2~e_3}$.
But by the equations in \eqref{thm:autodiff-correct-fst:eqn1}
and the definitions of $\semdiff{-}$, $\ifelse{-}{-}{-}$, and $\eval{D-}$,
we have $\eval{D\ifelse{\gamma_1}{\gamma_2}{\gamma_3}}(v)
= \semdiff{\code{if}~(e_1>0)~e_2~e_3}v$
for all $v \in \bR^N$. Hence, the theorem holds in this case as well.
%
%\xr{The following two constructions on partitions are used in several places;
%  how about defining them and instantiating them without redoing them each time:
%  (1) coarser partition that is finer than two partitions
%  (2) reverse image of a partition by a function
%  (it might save space and make the arguments clearer)
%}
%\wl{For a part of this, I newly defined $\ifelse{-}{-}{-}$.}
%\xr{Thanks, this is similar to what I meant, it is nice.}
\qed
\end{myproof}
%% ========== CONFN ==========
}

\showConfn{
%% ========== CONFN ==========
%% \begin{corollary}
%%         \label{cor:autodiff-correctness-fst}
{\bf \cref{cor:autodiff-correctness-fst}.\,}
{\it
        Assume that an autodiff system for the language in \S\ref{sec:autodiff} satisfies the two requirements in \S\ref{sec:autodiff}.
        Then, for each program $e$, there exists an intensional derivative $\mathit{df}$ in $\ideriv{\sem{e}}$
        such that if the system performs forward-mode (or reverse-mode) autodiff
        with a tangent vector $w \in \bR^N$ (or a cotangent vector $u \in \bR$),
        it computes the Jacobian-vector product $\mathit{df}(v) \cdot w \in \bR$
        (or the vector-Jacobian product $u^{\top} \!\!\cdot \mathit{df}(v) \in \bR^{N}$) %% 1 \times N
        for every input $v \in \bR^N$.
        Furthermore, the computed entity is
        the corresponding Jacobian-vector product (or vector-Jacobian product)
        with the standard derivative of $\sem{e}$ for almost all inputs $v \in \bR^N$.
}
%% \end{corollary}
\begin{myproof}
  Suppose that the two requirements are met, and let $e$ be a program.
  Consider the case when the system performs forward-mode autodiff.
  Let $\mathit{df} = \semdiff{e}$.
  Then, by Theorem~\ref{thm:autodiff-correct-fst},
  $\mathit{df}$ is an intensional derivative in $\ideriv{\sem{e}}$.
  Moreover, by the second requirement,
  the output of the system for $e$ with a tangent vector $w \in \bR^N$
  is $(\semdiff{e}v) \cdot w = \mathit{df}(v) \cdot w \in \bR$ for all inputs $v \in \bR^N$.
  This proves the first part of the corollary.
  The other part of the corollary follows immediately from \cref{prop:intr-deriv-coincidence}.
  The proof for the case when reverse-mode autodiff is performed is essentially the same,
  so we omit it.
  \qed
\end{myproof}
%% ========== CONFN ==========
}

\end{document}